 \documentclass[final,3p,times,twocolumn]{tempcls}

\usepackage{amssymb}
\usepackage{graphicx}
\usepackage{subfigure}
\usepackage{epstopdf}
\usepackage{multirow}
\usepackage{amsmath,amssymb,amsfonts}
\usepackage{amsopn}
\usepackage{graphicx}
\usepackage{subfigure}
\usepackage{epstopdf}
\usepackage{multirow}
\usepackage{array}
\usepackage{algorithm}
\usepackage{algorithmic}
\usepackage{arydshln}
\usepackage{amsthm}

\begin{document}

\begin{frontmatter}



\title{Shape Primitive Histogram: A Novel Low-Level Face Representation for Face Recognition}


\author[a]{Sheng Huang}
\author[a,b]{Dan Yang\corref{cor1}}

\author[d]{Haopeng Zhang}
\author[c]{Luwen Huangfu}

\author[b,e]{Xiaohong Zhang}
\address[a]{College of Computer Science at Chongqing University,400044,Chongqing,P.R.C}
\address[b]{School of Software Engineering at Chongqing Univeristy,400044,Chongqing,P.R.C}
\address[d]{School of Astronautics at Beihang University, 100191, Beijing, P.R.C}
\address[c]{State Key Laboratory of Management and Control for Complex Systems \\
Institute of Automation, Chinese Academy of Sciences, 100190,Beijing, P.R.C}

\address[e]{Ministry of Education Key Laboratory of Dependable Service Computing in Cyber Physical Society, 400044, Chongqing, P.R.C}
\cortext[cor1]{Corresponding author (Dan Yang): dyang@cqu.edu.cn}


\begin{abstract}
Human face contains abundant shape features. This fact motivates a lot of impressive shape feature-based face detection and 3D face recognition approaches. However, as far as we know, there is no prior low-level face representation which is purely based on shape feature proposed for conventional 2D (image-based) face recognition. In this paper, we present a novel low-level shape-based face representation named \textit{Shape Primitives Histogram} (SPH) for face recognition. In this approach, the face images are separated into a number of tiny shape fragments and we reduce these shape fragments to several uniform atomic shape patterns called \textit{Shape Primitives}. Then the face representation is obtained by implementing a histogram statistic of shape primitives in a local image region. In order to take scale information into consideration, we also produce Multi-scale Shape Primitive Histograms (MSPH) by concatenating the SPHs extracted from different scales. Moreover, we experimentally study the influences of each stage of SPH computation on performance, concluding that a small cell with 1/2 overlap and a fine size block with 1/2 overlap are important for good results. Four popular face databases, namely ORL, AR, YaleB and LFW-a databases, are employed to evaluate SPH and MSPH. Surprisingly, such seemingly naive shape-based face representations outperform the state-of-the-art low-level face representations.
%
\end{abstract}


\end{frontmatter}


\section{Introduction}
Face recognition is a fundamental task in biometrics, which is widely applied in our life.  As the core of face recognition, the quality of the face representation is the key for improving face recognition performance, since it is generally considered that the representation always determines the upper limit of the classification accuracy.

Many researchers have made efforts to find the effective face representations in recent decades. Generally speaking, the face representation approaches can be roughly categorized into two classes. The first one is called appearance-based approach which uses multivariate statistic analysis technique to learn a specific subspace from original high-dimensional sample space. This approach may start with the influential Eigenface~\cite{pca} and has produced many classical methods such as Fisherface~\cite{lda} and Laplacianface~\cite{lpp,lap,glpp,dhlp}. Moreover, this approach is also known as a dimensionality reduction technique for data analysis perspective. The second one is the low-level image feature-based face representation. This approach utilizes the pattern among the local pixels to represent and distinguish the faces. Garbor feature~\cite{gabor,gabor2,mrg} and Gradient feature~\cite{grad,gom} are deemed as two most common adopted low-level face representations. Both Gabor and gradient features are good at capturing the edge information of faces. Although extensive studies have proved their effectiveness, these methods are sensitive to the noise and local geometric transformation. In order to address the previous issues, some researchers develop a popular branch of low-level image representation named local histogram descriptor~\cite{hog,hogf,part,lbp,sift,learn,sadtf}. This approach will do the histogram statistics of the low-level image feature in a local image regions after the low-level image feature extraction. The main merit of local histogram descriptor over the conventional low-level image feature is that it is more insensitive to the local geometrical transformations and noises. Currently, Local Binary Patterns (LBP)~\cite{lbp,lbp2,lbp3,fsd} and Histogram of Oriented Gradient (HOG) \cite{gom,hog,hogf,part,hogf2} are the two most influential local histogram descriptors for face recognition. LBP exploits the local binary pattern among the pixels in a local circular region of the image. It is originally designed for texture description~\cite{lbp}. HOG is used to exploit the gradient orientation patterns in the image. Before being used as a face representation, it is known as a successful human detection feature~\cite{hog}. Although so many impressive low-level face representation methods have been proposed, most of those methods are based on gradient or edge information. However, human face contains abundant shape feature and the shape features are widely used in face detection~\cite{haar1,haar2,haar3,haar4} and 3D face recognition~\cite{lsdb,shape3f,shf,sfc}. So, in this paper, we intend to present a low-level face representation purely based on shape features for conventional 2D face recognition.

A popular way for shape feature extraction is Haar wavelet. Such approach has been demonstrated to be considerably successful in object detection field, especially in the face detection field~\cite{haar1,haar3,haar4}. This is mainly due to two facts. The first one is that human face contains abundant static shape characteristics~\cite{haar1,shape3f}. The second one is Haar feature, which provides an effective way to extract shape characteristics. Consequently, it can provide an attractive trade-off between accuracy and detection speed. The Haar feature based object detection has been in vogue for the past decade or so. Many impressive object detection systems and new Haar wavelet features have been proposed \cite{haar1,haar2,haar3,haar4}. However, as far as we know, there is no prior work that studies the representational power of Haar feature in the face recognition area. In this paper, we further exploit the representational power of Haar wavelet features to extract shape features for solving 2D face recognition task .
\begin{figure}[!tb]
\centering
\subfigure[]{
\centering
\includegraphics[scale=2.25]{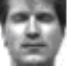}
\label{fig:subfig:a}}
\hspace{0.1in}
\subfigure[]{
\centering
\includegraphics[scale=0.5]{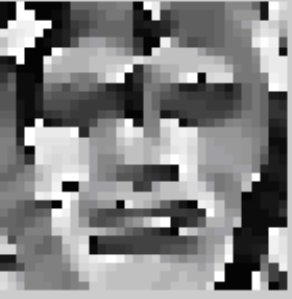}
}
\hspace{0.1in}
\subfigure[\mbox{}]{
\centering
\includegraphics[scale=0.52]{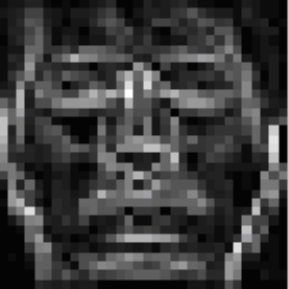}
}
\caption{The faces represented by Pixels,Gradient,Shape Primitive respectively}
\label{fig1}
\end{figure}

There exist extensive Haar wavelet templates. It is impracticable to apply all Haar templates for face representation. Instead, we consider that all the Haar wavelet templates that can be approximated by 14 square Haar wavelet templates and one flat template (see Figure \ref{fig3}). These 15 templates correspond to 15 atomic local shape characteristics (patterns). We call these atomic local shape patterns \textit{shape primitives} (see Figure \ref{fig1}). However, if we simply follow the same representation way as method of Viola and Jones \cite{haar4}, the face representation will have an incredibly high dimensionality and suffer more sensitivity from the local noises and geometrical transformations. Moreover, the shape characteristics which can benefit face recognition should be more local and detailed than the ones utilized by the face detection. So we adopt the form of local histogram descriptor to manage and describe the face shape features extracted by the shape primitives. Finally, as other histogram descriptors, we will get a vector by concatenating histograms of all the local image blocks. We name this new image descriptor \textit{Shape Primitive Histogram} (SPH). ORL face database is employed to experimentally learn the optimal parameters of SPH and Three larger face databases, namely AR, Yale-B and LFW-a face databases, are employed for evaluating the different face representations. Extensive experimental results demonstrate that our proposed method outperforms the state-of-the-arts.

The rest of paper is organized as follows: Section \ref{method} presents the generations of SPH and its multi-scale version; Section \ref{exp} describes the experiments for evaluating the SPH; Finally, the conclusion is summarized in Section \ref{con}.

\section{Methodology}
\label{method}
\begin{figure}[!htb]
  \centering
  \includegraphics[scale=0.5]{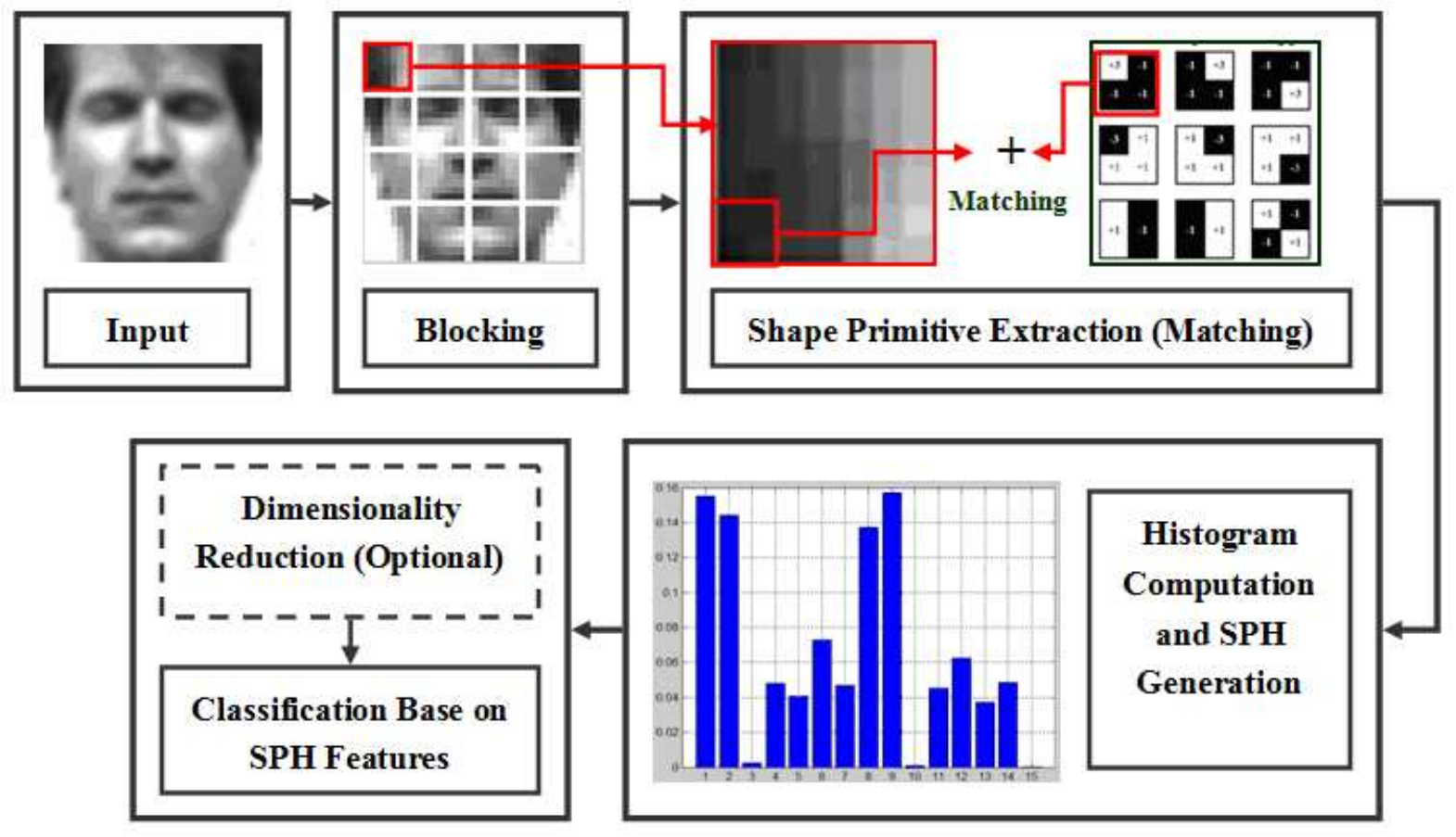}\\
  \caption{the overview of shape primitive histogram based face recognition system}
  \label{fig2}
\end{figure}

This section introduces shape primitive histogram in detail. The generation procedure of shape primitive histogram can be divided into three main steps: image blocking, shape primitive extraction (matching) and histogram computation. Figure \ref{fig2} depicts a face recognition system based on shape primitive histogram. After getting the SPH features, a dimensionality reduction can be employed for obtaining a more compact representation. And finally, the classifier is implemented for recognizing the faces (see Figure \ref{fig2}).

\subsection{Image Blocking}
 In order to incorporate the local spatial information, we divide the entire image into several same size blocks (local image patches, see the illustration of the second step in Figure \ref{fig2}). These blocks are the smallest unit for histogram statistics of shape primitives and determines the locality of SPH. Therefore, Shape primitive histogram is a local descriptor and the size of block can affect its performance. More specifically, if the block size is too small, the representation will be very sparse and thus leads to higher dimensionality. On the contrary, if its size is too big, the representation will be too rough to capture the local shape characteristics. In our case, the block is square. Moreover, the adjacent blocks are overlapping with each other. This strategy can reduce sensitivity to the geometric and photometric transformations~\cite{hog}.
\subsection{Shape Primitives Extraction (Matching)}
\begin{figure}[!htb]
  \centering
  \includegraphics[scale=0.5]{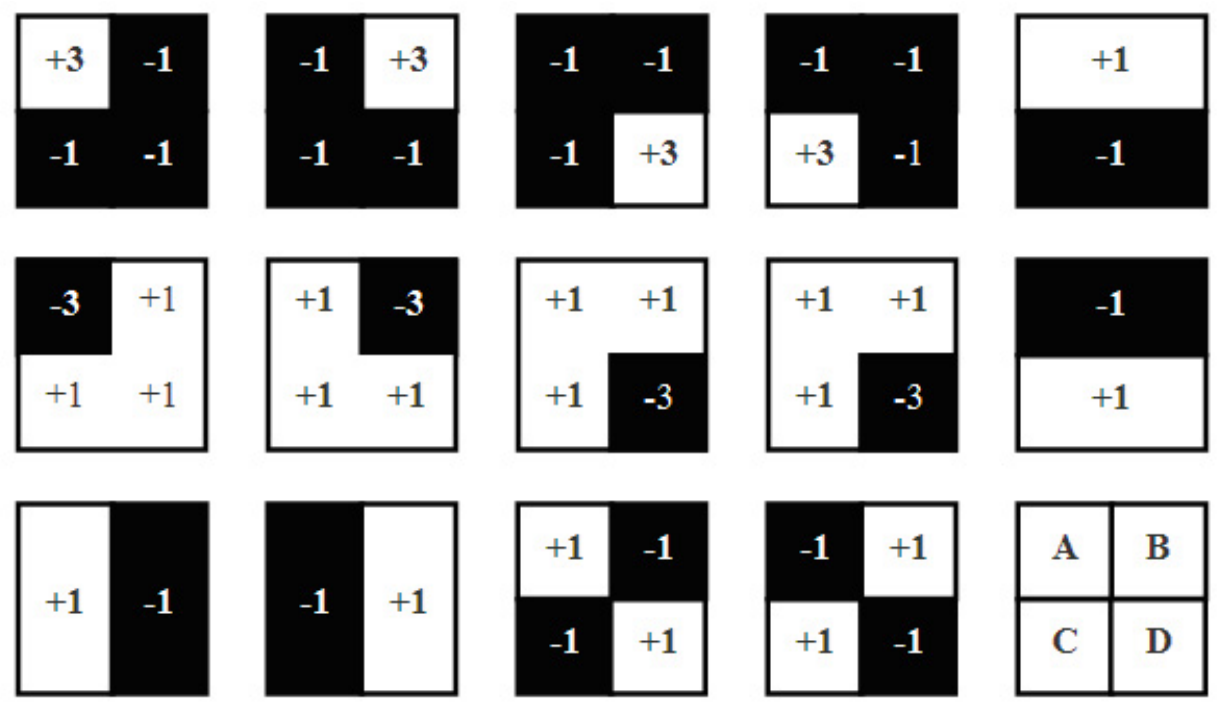}\\
  \caption{The Shape Primitives corresponding to the Haar wavelet templates.}
  \label{fig3}
\end{figure}

After image blocking, the second step is to extract shape primitives in each obtained image block. No matter how complex a shape is, it can be composed by 15 shape primitives. These shape primitives can be represented by the Haar wavelet templates in Figure~\ref{fig3}. The first 14 templates called non-flat shape primitive are used to describe the shape characteristics and the last template is a virtual template named flat shape primitive. The flat shape primitive is applied to handle the case that does't exist shape information.

As same as image blocking procedure, each local image block is divided into dozens of tiny $4^{n}$ pixels square fragments. Each fragment is an unit for extracting the shape primitive features. For simplicity, we name such fragment \emph{Cell}. As same as the templates in Figure \ref{fig3}, the cell has four bins. Each bin can be a pixel or a tiny square area which contains $2^n$ pixels. In order to keep continuity of shape information in a local block, each cell also has 1/2 overlap with the neighbor ones. Generally speaking, the size of cell determines the fineness of extracted local shape patterns. A small size of cell can capture more detailed shape information but also more sensitive to the noise. In our case, the cell size is fixed to 2$\times$2 pixels and 4$\times$4 pixels, since the size of face image is small.

The convolution operations based on  14 non-flat shape primitive templates are applied to each cell to find its belonging shape pattern. During procedure, 14 matching scores corresponding to 14 non-flat shape primitives are generated as follows:
\begin{equation}
S=\frac{1}{n}\sum_{i=1}^4 P_i h_i, \quad n=\max{(|h_1|,\cdots,|h_4|)}
\end{equation}
where $P_i$ indicates the sum of gray values of bin $i$ and $h_i$ indicates the weighting value corresponding to the bin $i$. The matching score represents the similarity between the cell and the shape primitive. Therefore, the cell must belong to the shape primitive which owns the maximum matching score, i.e.
\begin{equation}
\hat{i}=\arg\max_i{(\mathcal{S})}, \quad \mathcal{S}=(S_1,\cdots,S_i,\cdots,S_{14})
\end{equation}
However, the cell may not contain any shape information. In this case, the pixels in the cell have the same gray value. So, the matching scores of 14 shape primitives are all zero. In such case, the cell is actually a flat and we assume it is accord with the virtual template known as flat shape primitive. However, in the natural images, there seldom exists the absolute flat. So, for handling this case, we assume that the cell belongs to the flat shape primitive when all the matching scores are smaller than the nonnegative loose factor $\epsilon$ ($\epsilon\geq 0$).
Finally, the shape primitive extraction (matching) scheme can be further expressed as:
\begin{equation}
\hat{i}=\left\{
\begin{array}{l l}
{argmin_i(\mathcal{S}),~ M>{\epsilon}}\\
{15,~ M \leq {\epsilon}}
\end{array}
\right.
s.t. \quad \epsilon \geq 0
\end{equation}
where $M=\max(\mathcal{S})$ is the maximum among the matching scores and $\hat{i}$ indicates the index of the matched shape primitive. Since each non-flat shape primitive can find a complementary shape primitive among these 14 shape primitives, $M$ is always a positive or equal to zero, $M\geq 0$.

\subsection{Histogram Computation}
The histogram of each block has 15 bins corresponding to the 15 shape primitives. We can calculate the matching scores of the first 14 non-flat shape primitives. The flat one cannot achieve a matching score via convolution directly. However, these scores are the weighted votes of the histogram and will be accumulated into related bins during the histogram computation. In order to address this issue, we provide an empirical way to calculate matching score corresponding to the flat. The matching score (the weighting vote) of flat, $S_{15}$, can be assigned as follows:
\begin{equation}
S_{15}=\left\{
\begin{array}{l l}
{\epsilon-M+1,~M\leq{\epsilon}}\\
{0,~M>{\epsilon}}
\end{array}
\right.
\end{equation}
by this way, each cell can get the weighted vote and the weighted histogram statistics becomes feasible. So, the whole histogram computation procedure is denoted as follows
  \begin{equation}
\mathcal{H}_{i}=\left\{
\begin{array}{l l}
{\mathcal{H}_{i}+\epsilon-M+1,~M\leq{\epsilon}},~ i=15\\
{\mathcal{H}_{i}+M,~M>{\epsilon},~i=\arg\min_i(\mathcal{S})}
\end{array}
\right.
\end{equation}
where $\mathcal{H}_{i}$ is the value of the $i$th bin of the histogram and its initial value is zero. After finishing the histogram computation in each local block, a linear normalization, as $\mathcal{H}=\mathcal{H}/\sqrt{\mathcal{H}^T\mathcal{H}}$, is presented to every block for improving robustness to variation in illumination. Next, all the histograms are concatenated into a 1-D feature vector and this is the SPH feature.

\subsection{Multi-Scale Shape Primitive Histogram}
The cell size and block size are very important parameters, because they determine the fineness and locality of extracted features. More specifically, the SPH feature extraction via a smaller block with smaller cells can better capture more detailed shape characteristics, but it is hard to capture the more global shape characteristics and is more sensitive to the local noise. However, the quality and scale of images in a practical application are always extremely varied. Thus, a fixed size SPH may not be suitable to the image in an uncontrolled environment. For this case, we provide a very simple way to incorporate the scale information with SPH to improve its robustness. We hypothesize that a scale-robust SPH can be obtained by concatenating all SPH vectors extracted at different scales together. We name this new feature Multi-Scale Shape Primitives Histogram (MSPH). Note that this combined way is not optimal since the SPH features from different scales have correlations.

\section{Experiments}
\label{exp}
In this section, we conduct several experiments to learn the optimal parameters of SPH on ORL~\cite{ORL} database, which is a smaller database in comparison with the other three databases. AR~\cite{AR}, YaleB~\cite{Yaleb}, LFW-a~\cite{lfwa} datasets are employed for evaluating the performances of SPH and MSPH in comparison with three state-of-the-art face representations, namely Gabor Feature~\cite{gabor}, Local Binary Patterns (LBP)~\cite{lbp2} and Histogram of Oriented Gradient (HOG)~\cite{hogf}. Principal Component Analysis (PCA)~\cite{pca} and Linear Discriminant Analysis (LDA)~\cite{lda} are used as dimensionality reduction algorithms. The conventional classifier, Nearest Neighbor Classifier (NNC) and two more advanced classifiers, namely  Collaborative Representation Classifier (CRC)~\cite{CRC} and Support Vector Machine (SVM)~\cite{svm}, are adopted for classification.

\subsection{Datasets}
\begin{enumerate}
  \item The ORL database contains 400 images from 40 subjects \cite{ORL}. Each subject has ten images acquired at different times. In this database, the subjects' facial expressions and facial details are varying. And these images are also taken with a tolerance for some tilting and rotation of the face of up to $20^\circ$. The size of the face image is 32$\times$32 pixels. Compared to the other three databases, this database is much smaller and we use it to learn the optimal parameters.
  \item The AR database consists of more than 4,000 images of 126 subjects \cite{AR}. The database characterizes divergence from ideal conditions by incorporating various facial expressions, luminance alterations, and occlusion modes. Following paper \cite{LR}, a subset contains 1680 images with 120 subjects are constructed in our experiment. All these images are 50$\times$40 pixels.
  \item The Extended YaleB database \cite{Yaleb} consists of 2,414 frontal face images of 38 subjects under various lighting conditions. In our experiment, all of these images are 32$\times$32 pixels.
  \item The LFW-a database \cite{lfwa}, which aims at studying the problem of the unconstrained face recognition, is considered as one of the most challenging databases, since it contains 13233 images with great variations in terms of lighting, pose, age, and even image quality. We cropped these images to 120$\times$120 pixels around their center and resize these images to 32$\times$32 pixels for computational efficiency.
\end{enumerate}

Note, in our experiments, all the images on these four databases are grayscale images. With regard to the color image, the SPH feature can be extracted from each channel and yield them together as the new SPH feature.

\begin{figure*}[tb!]
\centering
\subfigure[ LDA-based two-fold cross-validation]{
\centering
\includegraphics[scale=0.36]{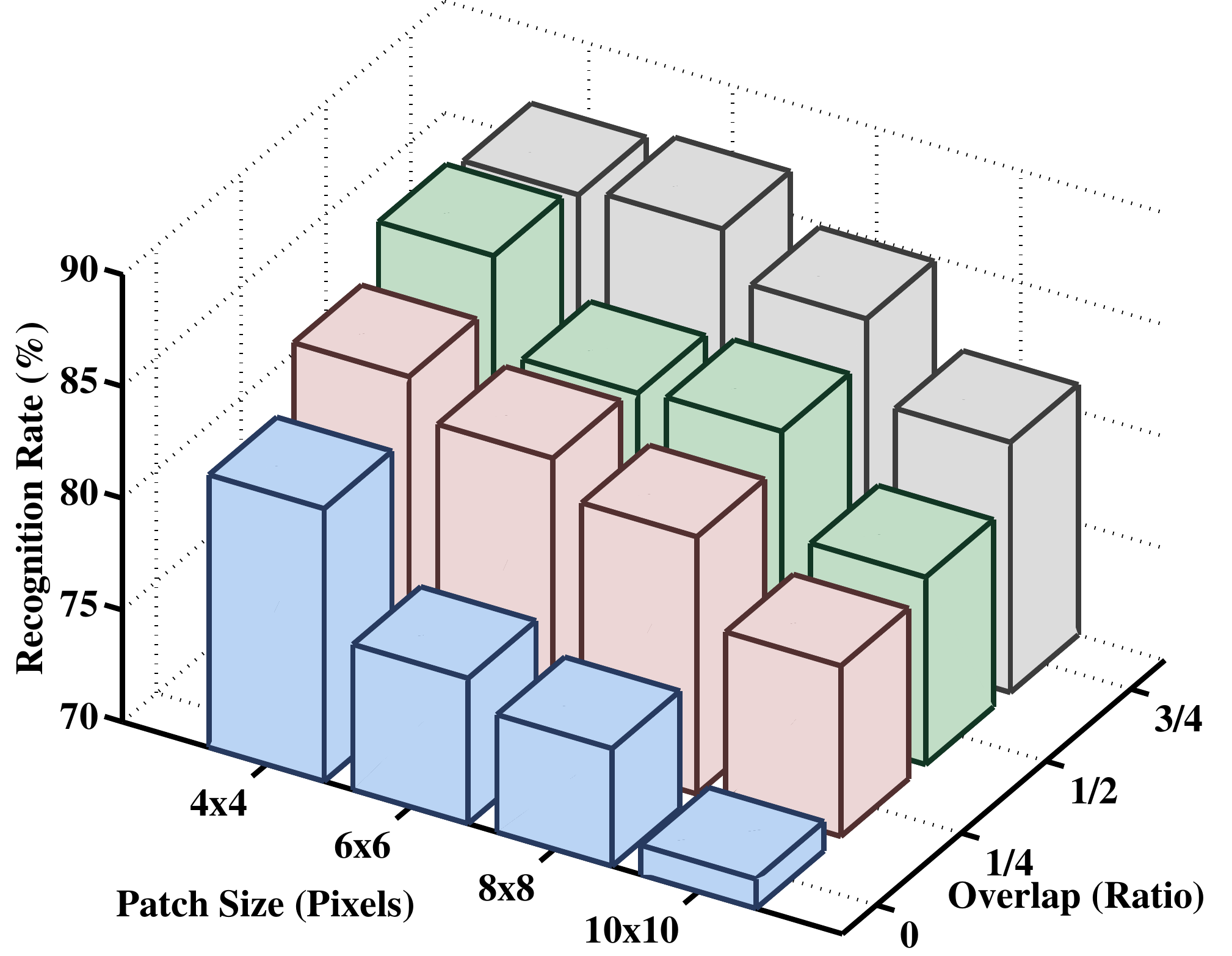}
\label{fig:subfig:a}}
\subfigure[PCA-based two-fold cross-validation ]{
\includegraphics[scale=0.36]{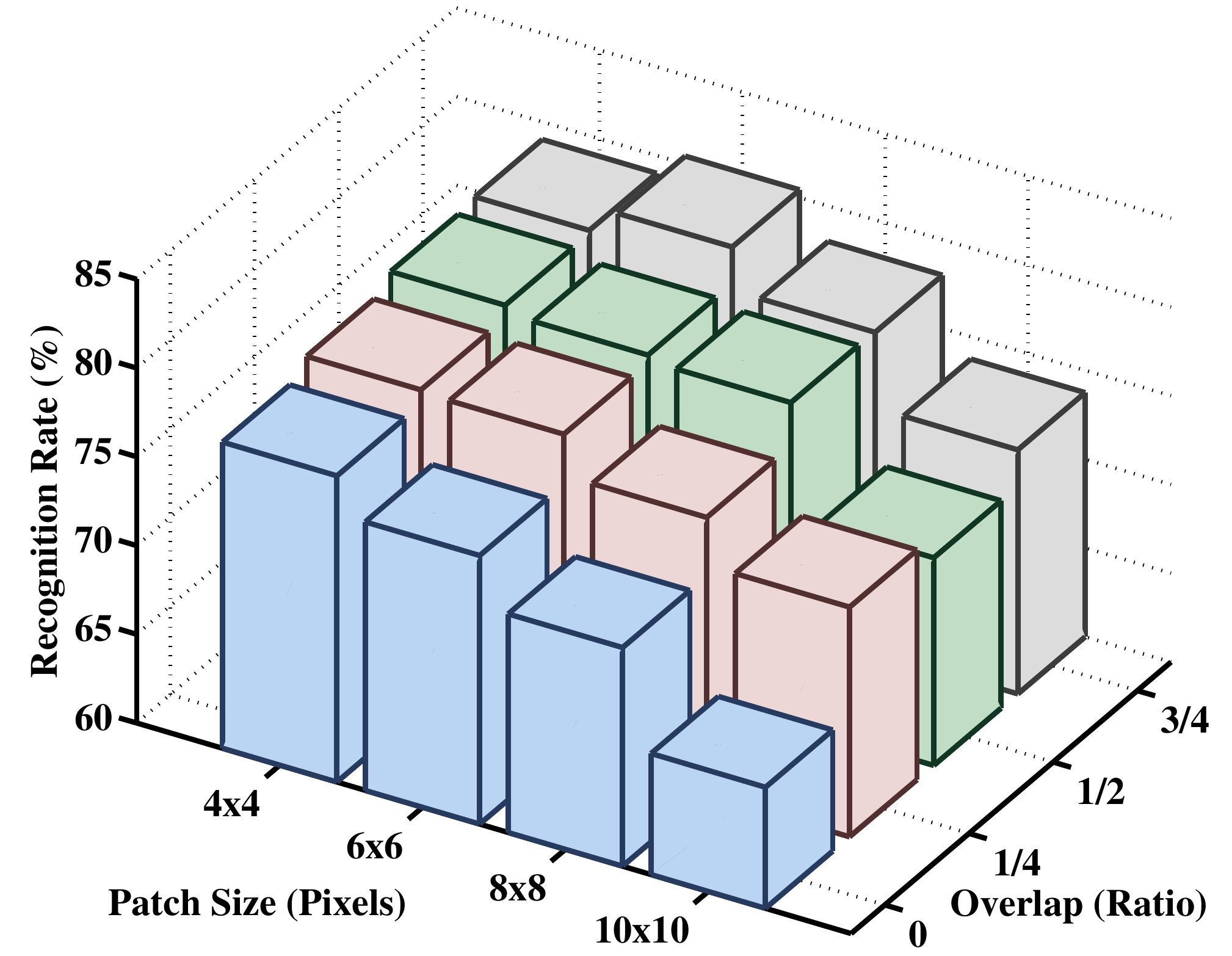}
}
\subfigure[LDA-based five-fold cross-validation ]{
\centering
\includegraphics[scale=0.36]{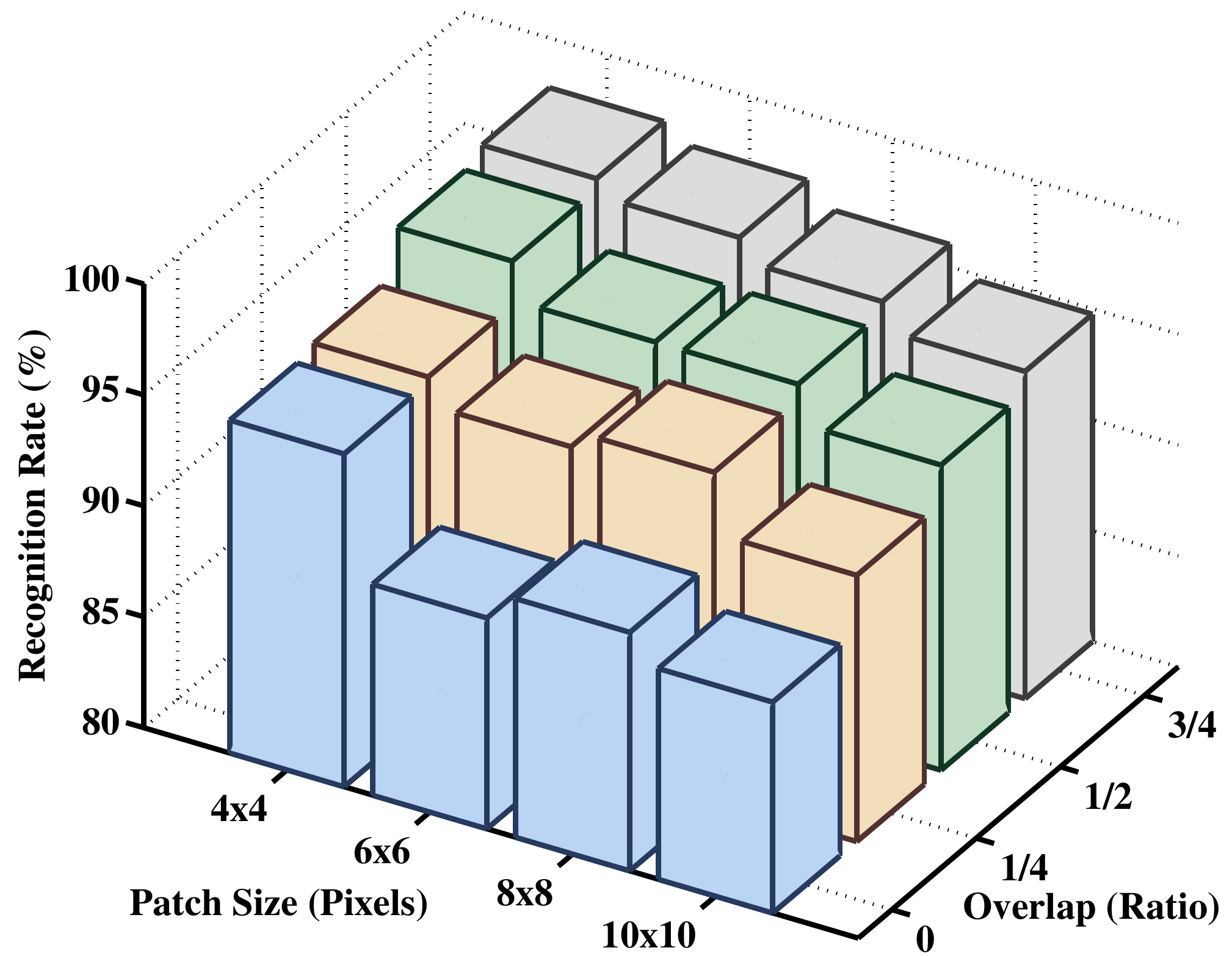}
}
\subfigure[ PCA-based five-fold cross-validation]{
\includegraphics[scale=0.36]{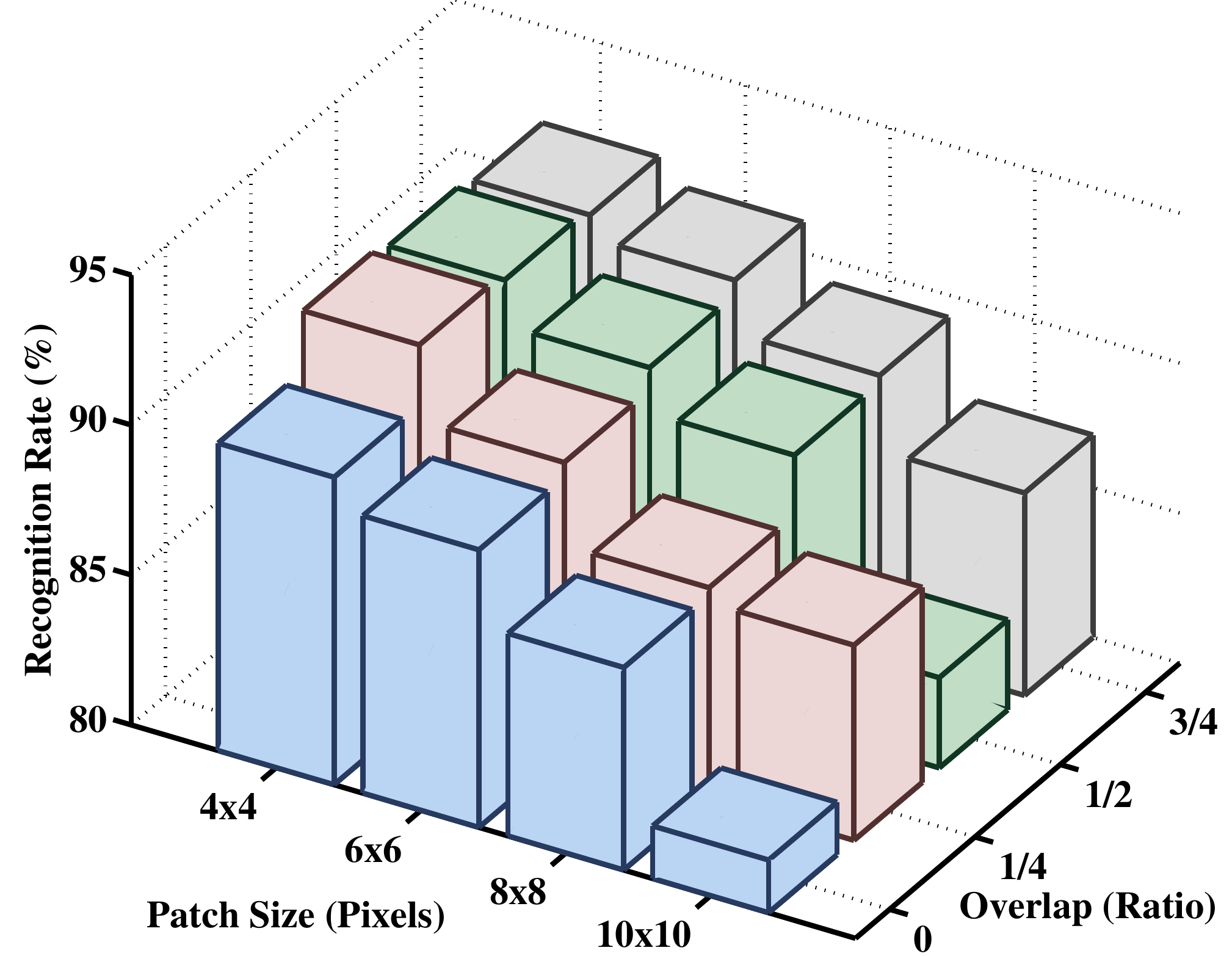}
}
\caption{The recognition rates under different the block sizes and overlaps. X axis indicates the overlapping area, Y axis indicates the block size and Z axis indicates the recognition rate.}
\label{fig4}
\end{figure*}
\begin{figure*}[tb!]
\centering
\subfigure[ ]{
\centering
\includegraphics[scale=0.38]{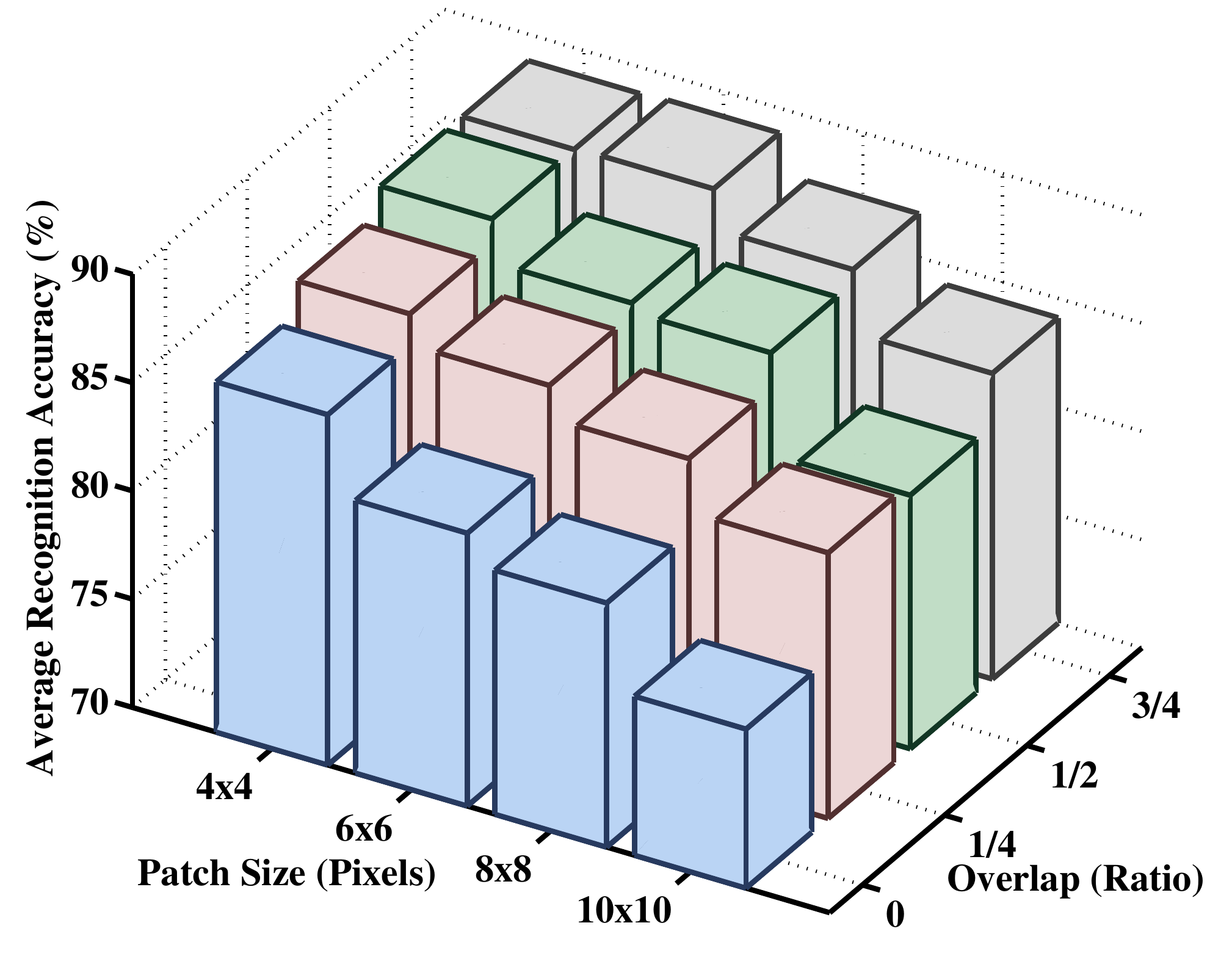}
\label{fig5a}}
\subfigure[ ]{
\centering
\includegraphics[scale=0.38]{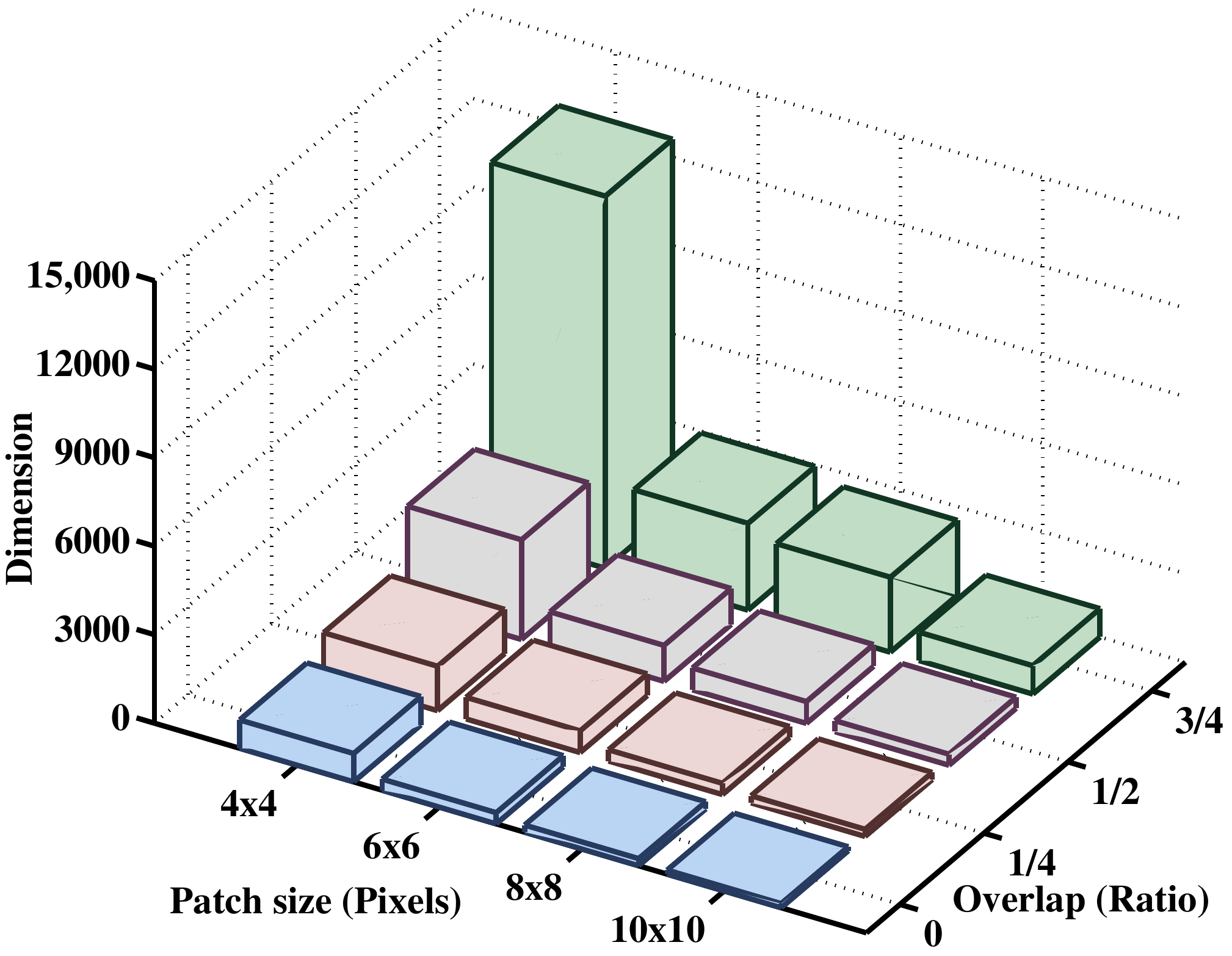}
\label{fig5b}}

\caption{(a) the comprehensive recognition rate, which is the mean of recognition rates under different cross-validations (see Figure \ref{fig4}). Y axis indicates block sizes, X axis indicates overlaps and Z indicates comprehensive recognition rates. (b) the dimensions of SPH under different block sizes and overlaps, Z indicates the dimension.}
\label{fig5}
\end{figure*}
\subsection{Parameters Learning}
SPH has several important parameters such as block size, overlap region, and loose factor which can influence the face recognition performance of SPH. Following \cite{hogf2}, the ORL database which is relatively smaller among four databases is utilized for experimentally learning the optimal values of these parameters. For a practical application, the training samples can be used for learning optimal parameters. Note, NNC is chosen as the classifier in these experiments. We adopt a learning strategy that fixes other parameters when one parameter is being experimentally learned.
\begin{figure}[!tb]
\centering
\includegraphics[scale=0.36]{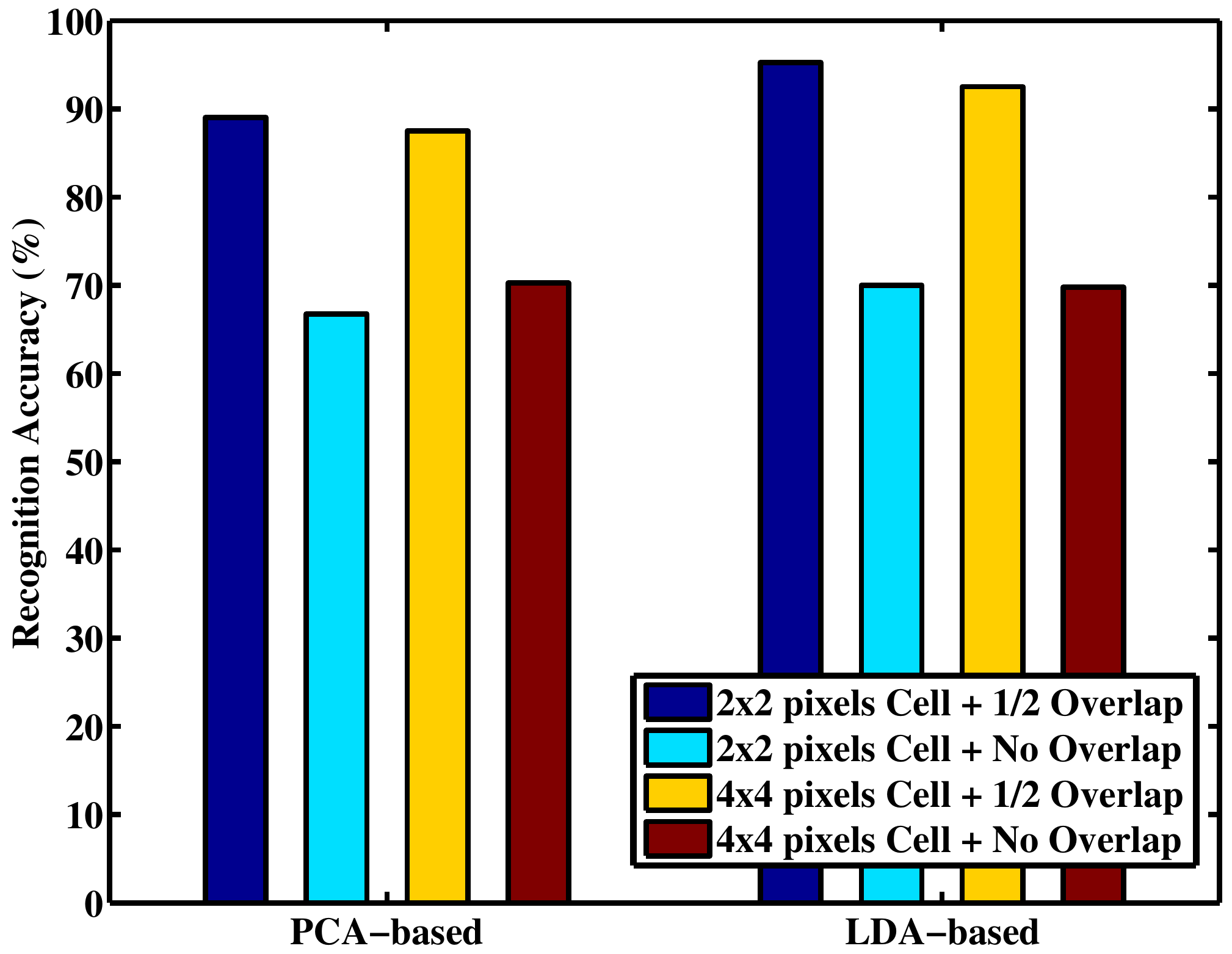}
\caption{The face recognition performances under different Cell Sizes and Overlaps.}
\label{Cell}
\end{figure}
\subsubsection{Sizes and Overlaps of Blocks}
In this experiment, four groups of blocks whose sizes are respectively $4\times{4}$, $6\times6$, $8\times8$, $10\times10$ pixels and four overlaps include no overlap, 1/4 overlap, 1/2 overlap, 3/4 overlap are adopted to produce a total of $4\times4=16$ combinations of blocks and overlaps for finding the best parameters of block and overlap. In this experiment, the cell size is fixed to 2$\times$2 pixels and LDA and PCA are applied for dimensionality reduction. Two-fold and five-fold cross-validations are employed for evaluating the recognition performance. The $n$-fold cross-validation in our paper is defined as: one part for training and the rest $n$-1 parts for testing.

Figure \ref{fig4} describes the influences of block size and overlap to the recognition performance on the ORL database. Figure \ref{fig5a} depicts the comprehensive recognition performance, which is the mean of recognition rates under different cross-validation schemes with different block sizes and overlaps. Figure \ref{fig5b} depicts the dimensions of different SPHs under different combinations of block size and overlap. From Figures \ref{fig4} and \ref{fig5}, it can be obviously concluded that SPH with a smaller block size and larger overlapping area can achieve higher recognition accuracy. However, the dimension of SPH also rapidly increases along with the block size reduction and the overlapping region expansion. In order to balance the recognition accuracy and recognition speed, a block size of 8$\times$8 pixels with 1/2 overlap are deemed as the optimal parameter group.

\begin{figure*}[htb!]
\centering
\subfigure[five-fold cross-validation ]{
\centering
\includegraphics[scale=0.376]{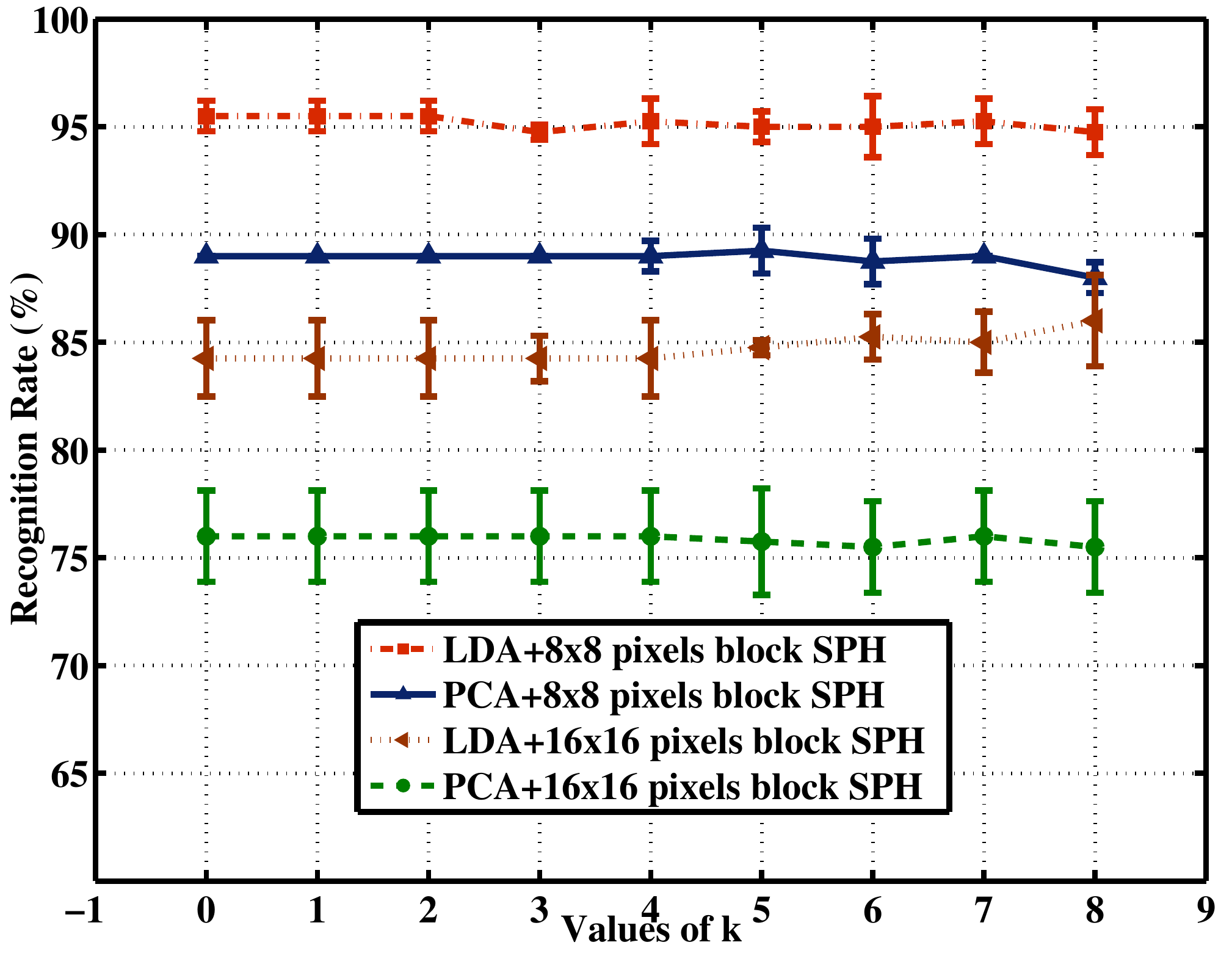}
\label{fig6a}}
\subfigure[ two-fold cross-validation]{
\includegraphics[scale=0.38]{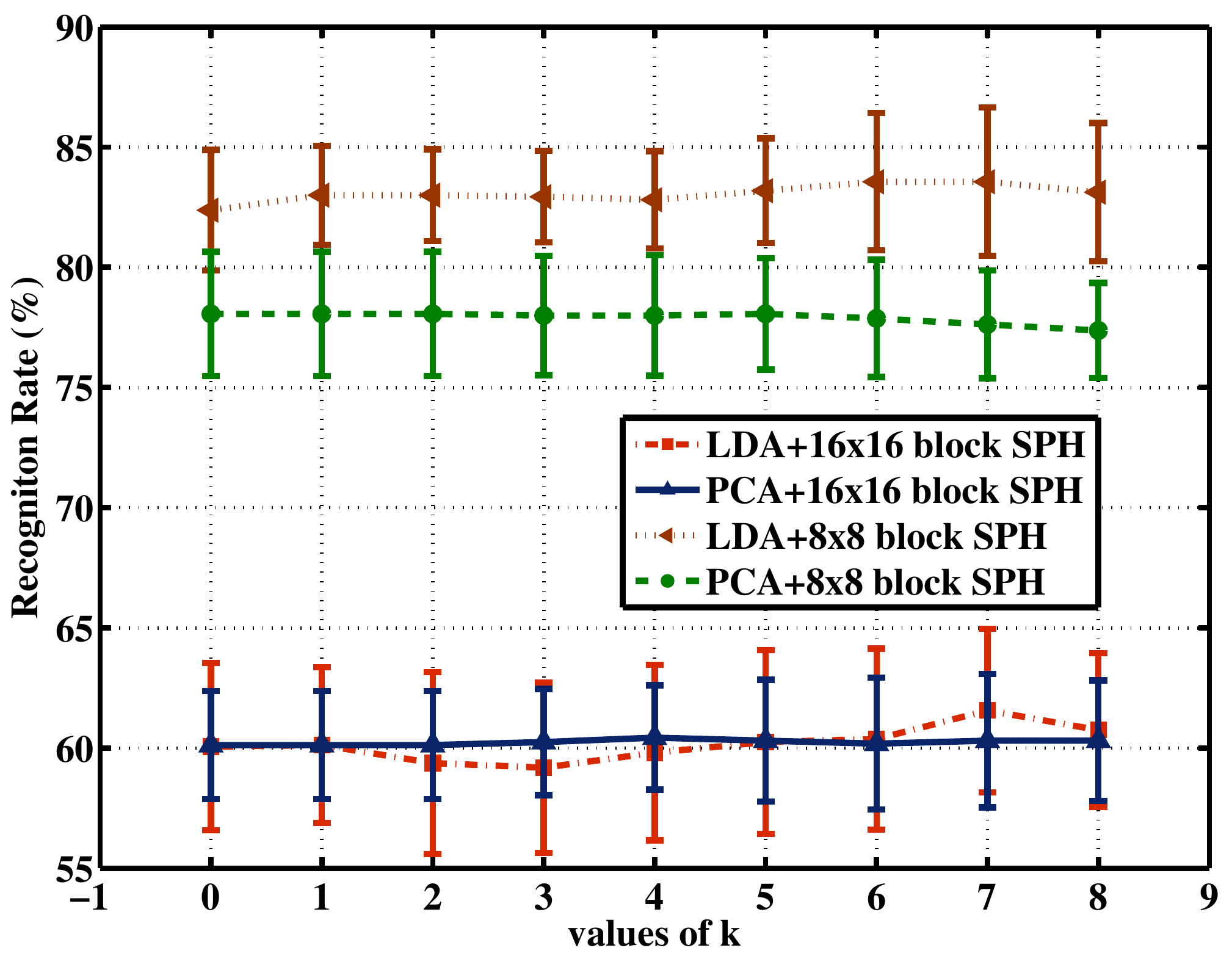}
\label{fig6b}}
\caption{ the recognition rates under different $k$. }
\label{fig6}
\end{figure*}
\subsubsection{Sizes and Overlaps of Cells}
Since the size of cell should be smaller than the size of block and the optimal block size is 8$\times$8 pixels, we choose two groups of cells whose sizes are respectively 2$\times$2 and 4$\times$4 pixels. Two overlaps include no overlap and 1/2 overlap are combined with these two groups of cells to produce four combinations. We apply two-fold cross-validation to evaluate the produced combinations. These experiments are all conducted under 8$\times$8 pixels block with 1/2 overlap. The results of experiments are shown in Figure \ref{Cell}.

From the observations, we can know that a smaller cell slightly outperforms a larger cell while the overlap in cells is a more important factor to improve the performance. The gain of the cell with 1/2 overlap over the cell without any overlap is around 20\%.  According to the results, we recommend to adopt the 2$\times$2 pixel size cell with 1/2 overlap for SPH extraction.

\subsubsection{The values of Loose Factor epsilon}
The loose factor $\epsilon$ controls the boundary between flat shape primitive and non-flat shape primitives and influence the flat shape primitive histogram weighting vote. So, it is very important to learn the optimal $\epsilon$. This parameter is related to the cell size $f$ (pixels), since the summation of the gray values of each bin of a bigger cell is greater. It can be denoted as follows:
\begin{equation}
\epsilon=k\times{(f/2)^2},k\ge{0}
\end{equation}
We can learn the value of $k$ to further achieve the optimal $\epsilon$ . In this experiment, two groups of SPH: 8$\times$8 pixels block with 1/2 overlap, 2$\times$2 pixels cell and 16$\times$16 pixels block with 1/2 overlap, 4$\times$4 pixels cell are used to study the effect of $\epsilon$. The two-fold and five-fold cross-validation schemes are adopted in these experiments.

According to the observations of Figure \ref{fig6}, which indicates the recognition accuracies under different $k$, we can conclude that two groups of SPH are all insensitive to $k$. Consequently, we let $k=1$. In other words, $\epsilon=f^{2}/4$ in this paper.

\begin{table}[htbp]
\footnotesize
  \begin{center}
    \begin{tabular}{c| c}
    \cline{1-2}
    Feature & Parameters Configurations \\ \hline
    SPH & 8$\times$8 pixels block, 1/2 overlap, 2$\times$2 pixels cell, $k=1$\\ \hline
    \multirow{3}{*}{MSPH} & 8$\times$8 pixels block, 1/2 overlap, 2$\times$2 pixels cell, $k=1$\\
    & 16$\times$16 pixels block, 1/2 overlap, 4$\times$4 pixels cell, $k=1$\\
    & 32$\times$32 pixels block, 1/2 overlap, 8$\times$8 pixels cell,$k=1$\\
    \hline

    \end{tabular}
    \caption{Parameters settings of SPH and MSPH in experiments}
    \label{t1}
  \end{center}
\end{table}

\begin{table*}[htb!]
\footnotesize
  \begin{center}
    \begin{tabular}{c l |c c| c c| c c}
    \hline
    \multicolumn{2}{c|}{ \multirow{3}*{Representation} }
    &\multicolumn{6}{c}{Cross-Validation Schemes-Recognition Rate {(\%)}}\\ \cline{3-8}
    & &\multicolumn{2}{c}{NNC}&\multicolumn{2}{c}{CRC}&\multicolumn{2}{c}{SVM}\\
    \cline{3-8}
    & & 7-fold & 4-fold& 7-fold & 4-fold & 7-fold & 4-fold \\
    \hline
        \multirow{5}*{PCA}  & \textbf{SPH}      &82.06$\pm$4.5\%&84.51$\pm$5.6\%&83.73$\pm$5.3\%&86.93$\pm$6.4\%&\textbf{83.79$\pm$5.3\%}&\textbf{87.03$\pm$6.5\%}\\
                        & \textbf{MSPH}     &80.72$\pm$4.8\%&83.12$\pm$5.3\%&82.54$\pm$6.0\%&85.28$\pm$7.0\%&82.94$\pm$6.3\%&85.70$\pm$7.7\%\\
                        & LBP~\cite{lbp2}   &71.27$\pm$6.7\%&76.69$\pm$6.9\%&76.30$\pm$8.2\%&81.69$\pm$9.7\%&76.43$\pm$8.8\%&81.95$\pm$9.9\% \\
                        & Gabor~\cite{gabor} &76.17$\pm$5.9\%&78.16$\pm$7.5\%&77.00$\pm$7.9\%&79.89$\pm$10.2\%&77.24$\pm$8.6\%&80.15$\pm$11.2\%\\
                        & HOG~\cite{hogf}&80.30$\pm$4.4\%&83.69$\pm$4.3\%&81.34$\pm$5.7\%&85.59$\pm$6.3\%&81.21$\pm$5.9\%&85.02$\pm$6.8\%\\

                        \hdashline
    \multirow{5}*{LDA}  & \textbf{SPH}      &\textbf{83.16$\pm$6.0\%} &\textbf{86.67$\pm$7.3\%}&\textbf{84.56$\pm$5.2\%}&\textbf{88.28$\pm$6.1\%}&80.71$\pm$6.4\%&86.25$\pm$7.3\%\\
                        & \textbf{MSPH}     &79.29$\pm$7.1\% &85.01$\pm$8.2\%&81.84$\pm$6.1\%&84.92$\pm$8.4\%&80.39$\pm$7.2\%&84.32$\pm$8.6\% \\
                        & LBP~\cite{lbp2}   &74.38$\pm$8.4\% &82.48$\pm$8.0\%&76.67$\pm$8.4\%&82.63$\pm$8.5\%&76.40$\pm$8.4\%&83.01$\pm$8.5\% \\
                        & Gabor~\cite{gabor}   &79.74$\pm$7.2\% &83.30$\pm$10.2\%&79.59$\pm$7.7\%&84.09$\pm$9.5\%&77.84$\pm$8.2\%&83.54$\pm$10.0\%\\
                        & HOG~\cite{hogf}&75.18$\pm$6.8\% &80.66$\pm$8.0\%&75.59$\pm$7.6\%&75.02$\pm$10.3\%&75.11$\pm$8.0\%&74.94$\pm$10.3\%\\
                        \hline

    \end{tabular}
    \caption{Recognition accuracies on AR Database}
    \label{t2}
  \end{center}
\end{table*}

\begin{table*}[tb!]
\footnotesize
  \begin{center}
    \begin{tabular}{c l |c c| c c| c c}
    \hline
    \multicolumn{2}{c|}{ \multirow{3}*{Representation} }
    &\multicolumn{6}{c}{Leave-one-out Cross-Validation-Recognition Rate {(\%)}}\\ \cline{3-8}
    & &\multicolumn{2}{c}{NNC}&\multicolumn{2}{c}{CRC}&\multicolumn{2}{c}{SVM}\\
    \cline{3-8}
    & & 8-fold & 5-fold& 8-fold & 5-fold& 8-fold & 5-fold \\
    \hline
        \multirow{5}*{PCA}  & \textbf{SPH}      &52.51$\pm$11.2\%&60.00$\pm$3.2\%&\textbf{57.49$\pm$11.6\%}&\textbf{66.74$\pm$6.0\%}&\textbf{60.99$\pm$12.4\%}&\textbf{70.02$\pm$5.6\%}\\
                        & \textbf{MSPH}     &45.68$\pm$11.2\%&53.56$\pm$3.4\%&49.78$\pm$11.5\%&59.65$\pm$4.7\%&52.27$\pm$11.0\%&63.15$\pm$4.5\% \\
                        & LBP~\cite{lbp2}   &33.27$\pm$12.3\%&40.59$\pm$7.2\%&37.58$\pm$12.3\%&47.40$\pm$7.6\%&37.74$\pm$11.5\%&46.26$\pm$6.0\% \\
                        & Gabor~\cite{gabor} &43.30$\pm$9.1\%&51.63$\pm$7.5\%&49.99$\pm$12.3\%&58.70$\pm$6.2\%&48.22$\pm$8.3\%&57.53$\pm$6.0\% \\
                        & HOG~\cite{hogf}&39.31$\pm$11.2\%&46.64$\pm$2.2\%&43.33$\pm$10.6\%&51.98$\pm$3.9\%&43.00$\pm$8.3\%&51.28$\pm$3.4\%\\

                        \hdashline
    \multirow{5}*{LDA}  & \textbf{SPH}       &\textbf{58.88$\pm$9.3\%}&\textbf{66.65$\pm$6.0\%}&53.95$\pm$9.3\%&55.35$\pm$4.5\%&55.40$\pm$9.0\%&56.60$\pm$5.1\%\\
                        & \textbf{MSPH}     &55.67$\pm$9.0\%&62.58$\pm$6.7\%&53.18$\pm$9.0\%&57.25$\pm$5.7\%&54.24$\pm$9.0\%&57.04$\pm$5.1\%\\
                        & LBP~\cite{lbp2}    &41.03$\pm$11.2\%&49.11$\pm$3.4\%&37.47$\pm$10.3\%&35.42$\pm$4.6\%&38.05$\pm$10.5\%&35.01$\pm$4.8\% \\
                        & Gabor~\cite{gabor}   &50.45$\pm$4.5\%&57.29$\pm$6.4\%&48.04$\pm$6.9\%&53.98$\pm$2.5\%&38.05$\pm$10.5\%&35.01$\pm$4.8\% \\
                        & HOG~\cite{hogf}&39.31$\pm$11.2\%&48.03$\pm$3.6\%&32.06$\pm$8.3\%&31.21$\pm$3.6\%&32.55$\pm$7.9\%&31.31$\pm$3.8\%\\
                        \hline

    \end{tabular}
    \caption{Recognition accuracies on YaleB Database}
    \label{t3}
  \end{center}
\end{table*}
\begin{table*}[tb!]
\footnotesize
  \begin{center}
    \begin{tabular}{c l |c c| c c| c c}
    \hline
    \multicolumn{2}{c|}{ \multirow{3}*{Representation} }
    &\multicolumn{6}{c}{Leave-one-out Cross-Validation-Recognition Rate {(\%)}}\\ \cline{3-8}
    & &\multicolumn{2}{c}{NNC}&\multicolumn{2}{c}{CRC}&\multicolumn{2}{c}{SVM}\\
    \cline{3-8}
    & & subset1 & subset2 & subset1 & subset2 & subset1 & subset2 \\
    \hline
    \multirow{5}*{PCA}  & \textbf{SPH}      &19.05$\pm$2.5\%&19.83$\pm$3.6\%&25.62$\pm$5.6\%&37.65$\pm$4.4\%&26.53$\pm$2.5\%&40.09$\pm$4.1\%\\
                        & \textbf{MSPH}     &19.05$\pm$2.7\%&20.83$\pm$3.6\%&26.53$\pm$5.7\%&\textbf{38.23$\pm$4.2\%}&27.66$\pm$2.8\%&41.26$\pm$3.9\% \\
                        & LBP~\cite{lbp2}   &22.79$\pm$2.5\%&23.77$\pm$4.5\%&31.63$\pm$1.1\%&38.11$\pm$5.1\%&29.82$\pm$0.7\%&42.66$\pm$4.6\% \\
                        & Gabor~\cite{gabor}   &15.42$\pm$1.6\%&19.40$\pm$2.4\%&17.91$\pm$2.2\%&30.42$\pm$3.1\%&19.61$\pm$1.6\%&32.87$\pm$5.4\%\\
                        & HOG~\cite{hogf}&19.27$\pm$3.5\%&22.62$\pm$2.8\%&25.17$\pm$2.2\%&34.38$\pm$4.2\%&23.81$\pm$3.4\%&35.20$\pm$4.3\%\\
                        \hdashline
    \multirow{5}*{LDA}  & \textbf{SPH}      &24.94$\pm$3.5\%&35.00$\pm$4.3\%&29.71$\pm$3.4\%&37.65$\pm$4.2\%&30.84$\pm$3.4\%&44.38$\pm$3.6\%\\
                        & \textbf{MSPH}     &\textbf{27.66$\pm$2.8\%}&\textbf{35.79$\pm$3.6\%}&\textbf{32.09$\pm$2.7\%}&36.95$\pm$4.3\%&\textbf{32.31$\pm$2.7\%}&\textbf{44.60$\pm$4.4\%} \\
                        & LBP~\cite{lbp2}   &23.58$\pm$2.9\%&30.35$\pm$5.3\%&28.68$\pm$4.2\%&32.40$\pm$4.7\%&28.91$\pm$3.5\%&41.66$\pm$4.4\% \\
                        & Gabor~\cite{gabor}   &22.56$\pm$3.0\%&33.65$\pm$5.6\%&26.30$\pm$3.0\%&33.57$\pm$3.3\%&26.30$\pm$2.6\%&43.59$\pm$4.9\%\\
                        & HOG~\cite{hogf}&18.71$\pm$2.3\%&24.84$\pm$3.8\%&22.68$\pm$2.8\%&27.04$\pm$4.2\%&22.90$\pm$3.1\%&32.86$\pm$3.1\%\\
                        \hline

    \end{tabular}
    \caption{Recognition accuracies on LFW-a Database}
    \label{t4}
  \end{center}
\end{table*}
\subsection{Face Recognition}
Three larger face databases including AR, YaleB and LFW-a databases are employed for evaluating the face recognition performance. Among these three face databases, the LFW-a database is a face database in uncontrolled environment which is a very challenging database aiming at the evaluation of face recognition in the wild. The sample number of each subject in this database is very different. Following paper~\cite{rcr}, we divide the LFW-a database into two subsets. The first subset (147 subjects, 1100 samples) is constructed by the subjects whose sample numbers are ranged from 5 to 10 and the second subset (127 subjects, 2891 samples) is constructed by the subjects whose sample numbers are all over 11. We apply leave-one-out cross-evaluation scheme to these two subsets. With regard to AR database, we employ 7-fold and 4-fold cross-validation schemes. While, the cross-validation schemes of YaleB are 8-fold and 5-fold. The parameters configurations of SPH and MSPH in these experiments are introduced in Table \ref{t1}. The parameter settings of LBP, Gabor and HOG are mainly following \cite{lbp2}, \cite{gabor} and \cite{hogf} respectively. But the block sizes of HOG and LBP are slightly tuned for fitting the face image size in our experiments (the sizes of images in their experiments are quite different to us). The block sizes of LBP and HOG are all 16$\times$16 pixels and each block has 1/2 overlap with the neighbour ones.

\begin{figure*}[htb!]
\centering
\subfigure[Recognition accuracy in PCA space on AR database]{
\centering
\includegraphics[scale=0.36]{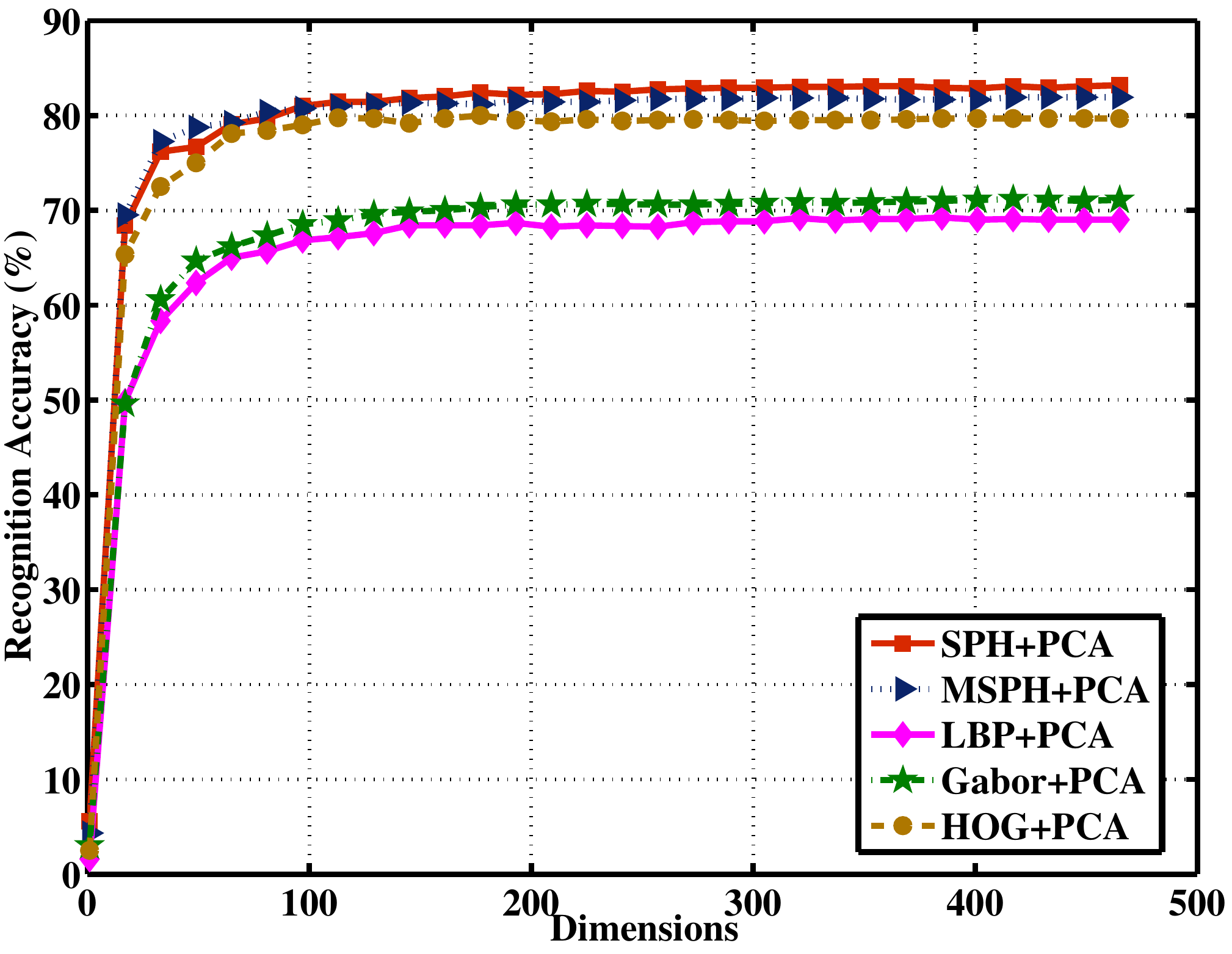}
\label{fig:subfig:a}}
\subfigure[Recognition accuracy in LDA space on AR database]{
\includegraphics[scale=0.36]{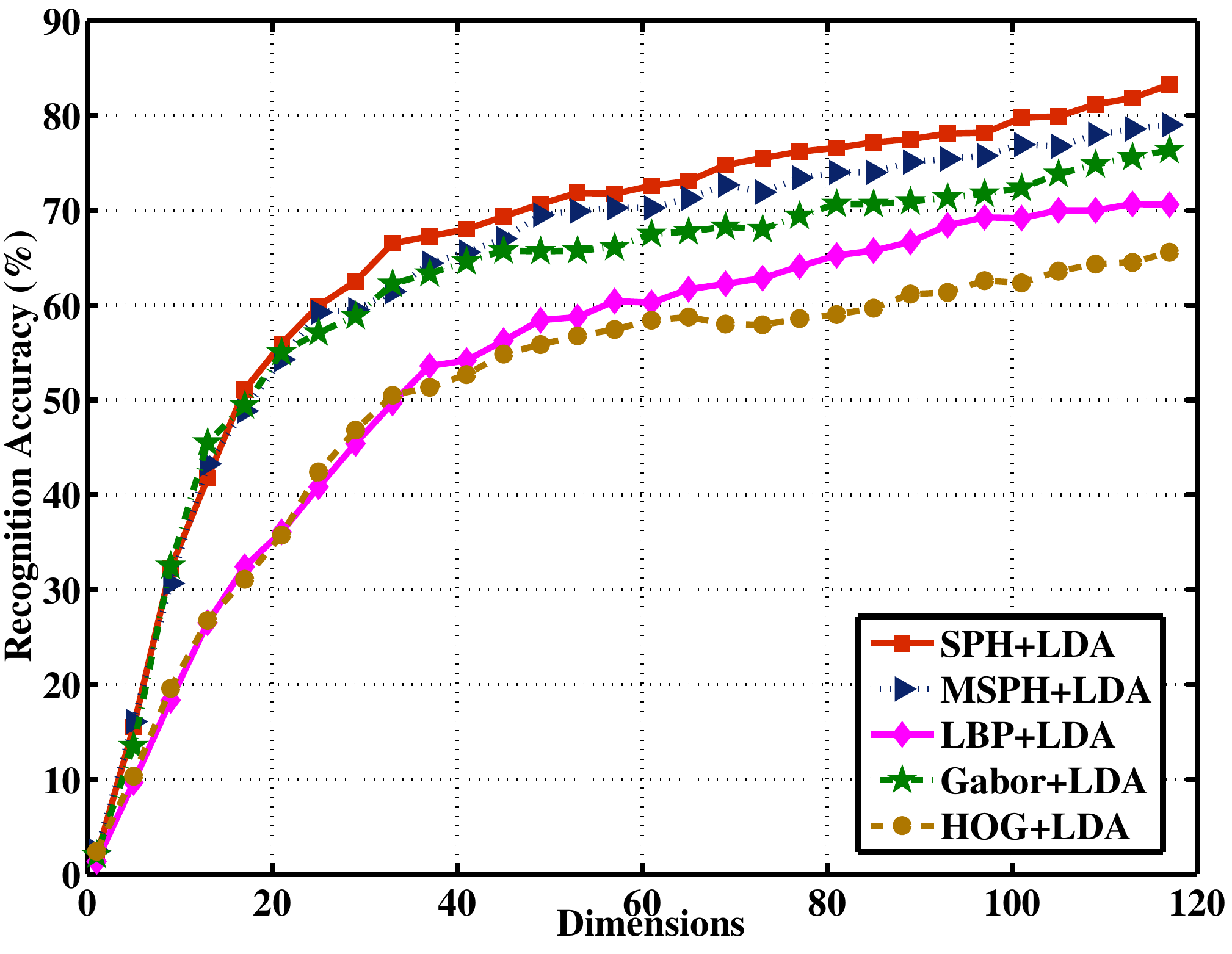}
}
\subfigure[Recognition accuracy in PCA space on YaleB database ]{
\centering
\includegraphics[scale=0.36]{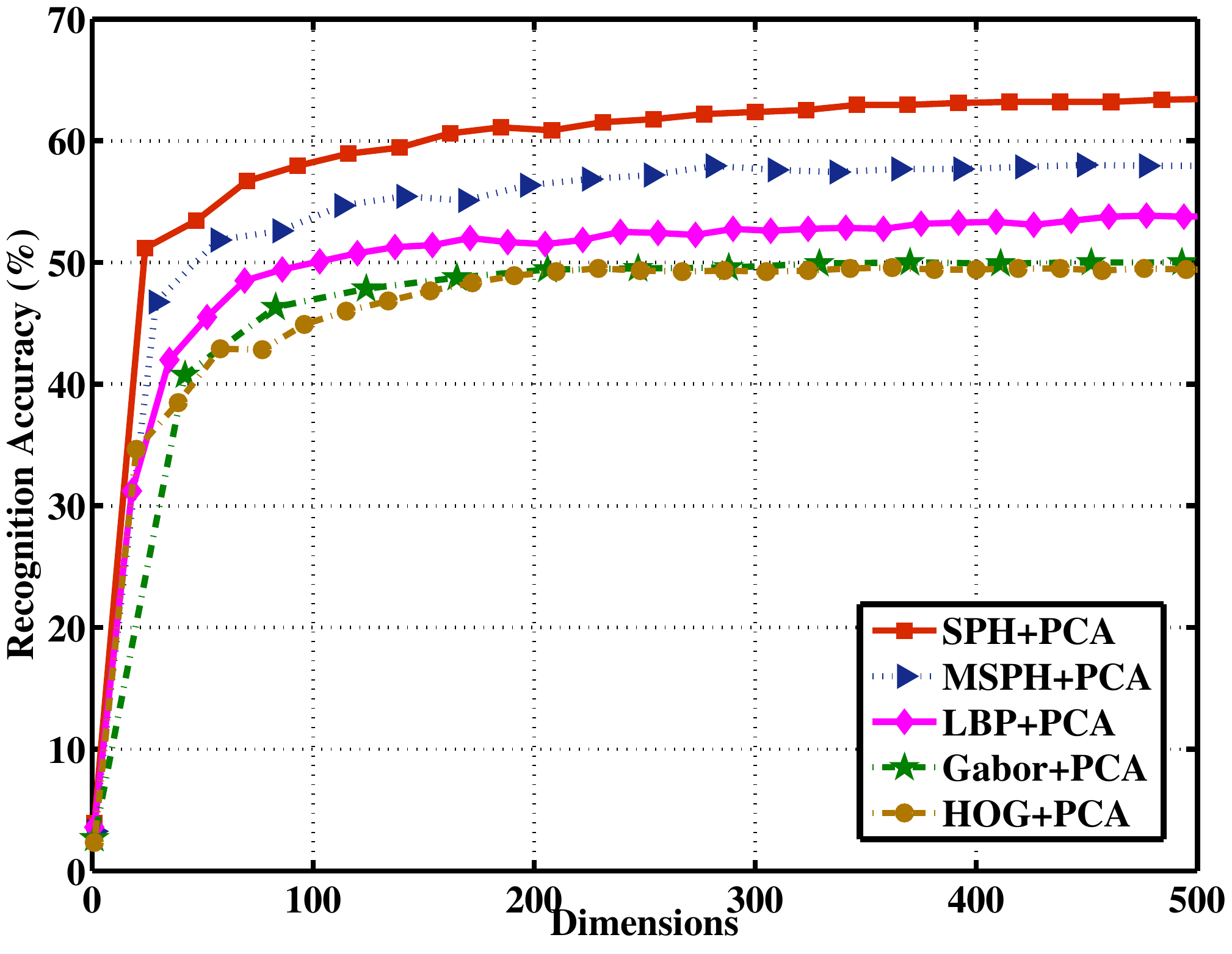}
}
\subfigure[Recognition accuracy in LDA space on YaleB database ]{
\includegraphics[scale=0.36]{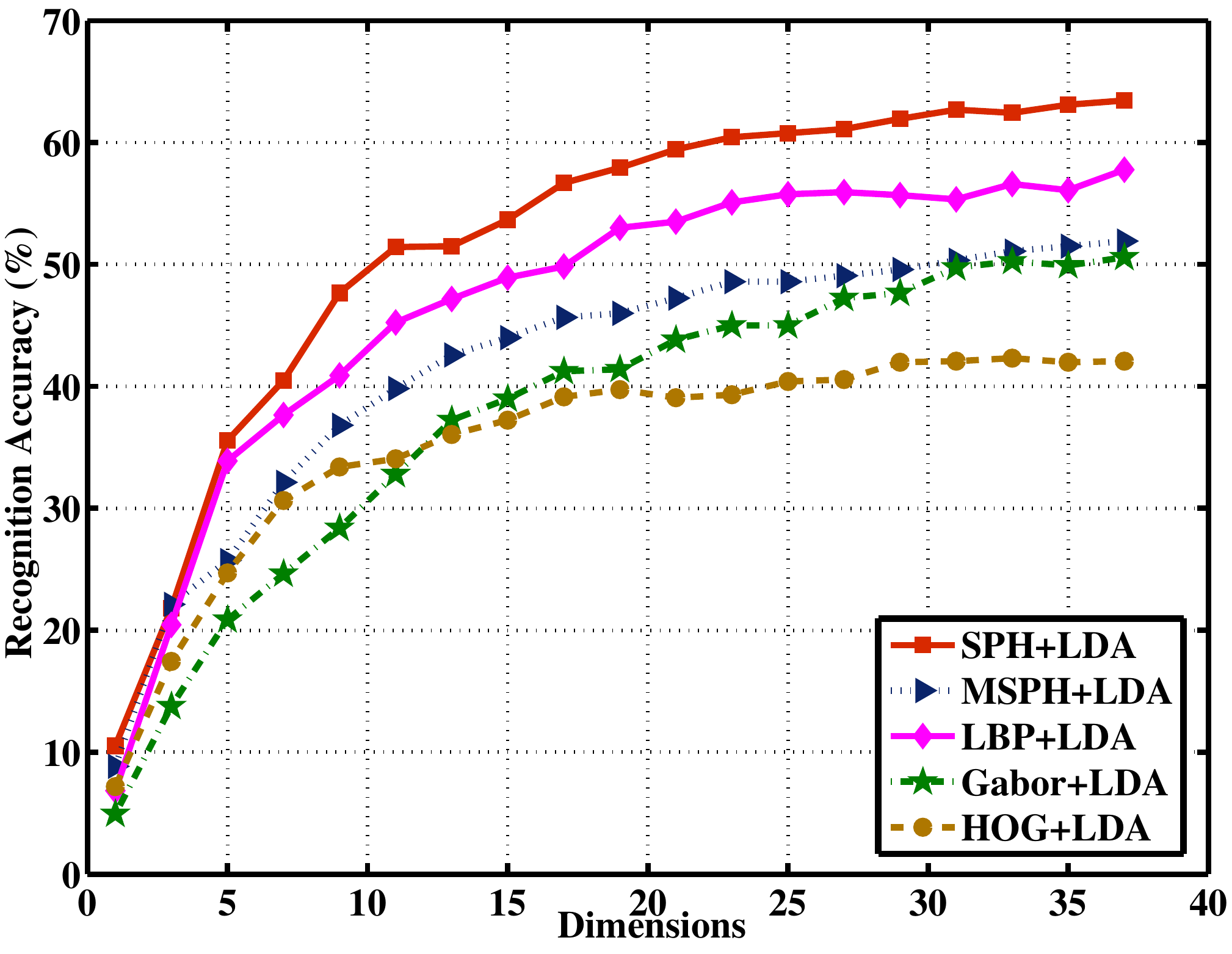}
}
\caption{Recognition Accuracy versus Retained Dimensions.}
\label{ravrd}
\end{figure*}

Tables \ref{t2}, \ref{t3} and \ref{t4} show the face recognition accuracies using different face representations on AR, YaleB and LFW-a databases respectively.  It is obvious that SPH outperforms the other three state-of-the-art representations on AR and YaleB databases under all three classifiers. For example, SPH obtains averagely 3.5\%, 4.5\% and 2.25\% gain over the second top representation using NNC, CRC and SVM classifiers respectively on AR database. On YaleB database, these numbers are respectively 8.25\%, 7.75\% and 12\%. Additionally, the MSPH can also maintain the second place on these two databases in the most of time. With regards to the experimental results on the LFW-a database, MSPH gets a better performance than SPH and defeats other compared methods under all three different classifiers. The reason why the MSPH gets a better performance than SPH using the LFW-a database may be the fact that the samples in the LFW-a database suffer from more variation in image resolution. Thus, the MSPH can perform better. Besides that, Table \ref{t4} demonstrates that SPH also obtains very promising performance.

For better demonstrating the superiority of our method and studying the influence of dimensionality reduction algorithms to the low-level face representations, we conduct several experiments on AR and YaleB databases to draw the retained dimensions versus recognition accuracies curves in Figure \ref{ravrd}. On AR database, the first 10 samples per subject are used for training while the rest samples are for testing. On YaleB database, the first 48 samples per subject are used for training while the rest samples for testing. From the observations in Figure \ref{ravrd}, we can know that SPH outperforms other compared methods in all dimensions while MSPH also outperforms the compared methods except in the experiments on YaleB database that uses LDA for dimensionality reduction. The results of experiments clearly demonstrate that SPH is more discriminative than the other three face representations.
\begin{figure}[htbp]
\centering
\includegraphics[scale=0.36]{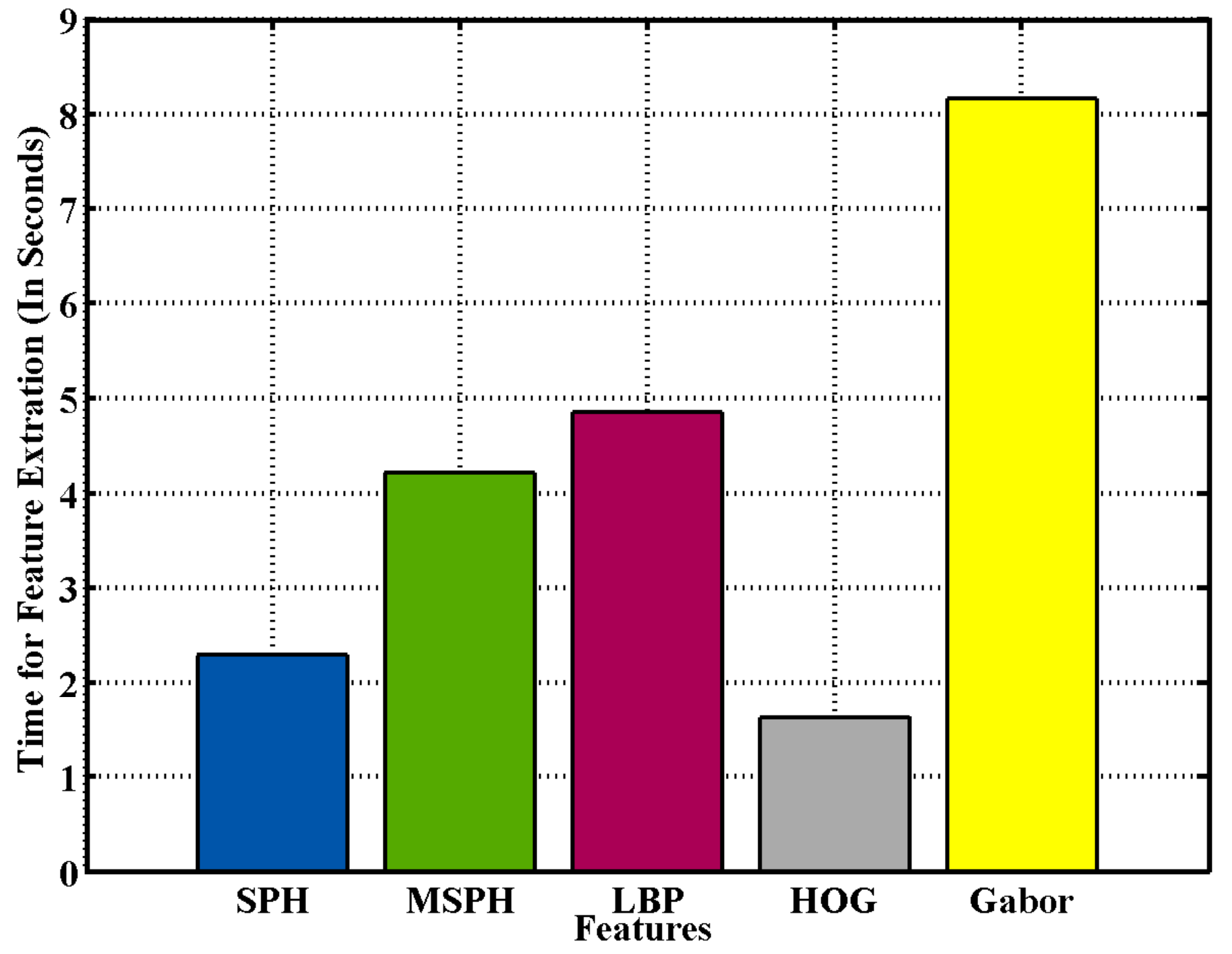}
\caption{The extraction time of different features on ORL database.}
\label{times}
\end{figure}

\subsection{Feature Extraction Efficiency}
For more comprehensively evaluating SPH, we also test the feature extraction efficiencies of different low-level face representations on ORL database and report the results in Figure \ref{times}. The experimental hardware configuration is CPU: 2.5 GHz, RAM: 8G. The results of experiments in Figure \ref{times} clearly show that feature extraction speeds of SPH and MSPH are competitive.

\section{Conclusion}
\label{con}
In this paper, we have proposed a simple but effective low-level face representation for face recognition. This representation focuses on highlighting the shape characteristics of face and supposes human face can be divided into a group of small shape fragments, which share a series of uniform shape patterns named \emph{Shape Primitives}. A histogram of shape primitives is computed in each local block and be concatenated as a 1-D vector to represent the face. Moreover, we have also generated a multi-scale shape primitive histogram via concatenating the different scale SPH vectors together. Three well known face databases were used for validating the proposed methods. The experimental results demonstrate the superiority of SPH in comparison with the state-of-the-art low-level face representation methods.

There are many worthwhile works of our method can be further exploited. For example, this descriptor can be combined with facial landmark localizations or face interesting region selections to increase the recognition accuracy \cite{hogf, fpatch,yupose}. Currently, face recognition via fusing different features is a very popular trend for face recognition \cite{fusion}. So, we can also fuse SPH with other state-of-the-art low-level features to present a more powerful representation. Moreover, applying SPH to face expression analysis, face alignment and face spoofing detection~\cite{fsd} are also interesting directions.
\section*{Acknowledgement}
The work described in this paper was partially supported by National Natural Science Foundations of China (NO. 60975015 and 61173131), Fundamental Research Funds for the Central Universities (No. CDJXS11181162). The authors would like to thank the helpful suggestions from Mr. Mark Dilsizian and the useful comments of the anonymous reviewers and editors.




\bibliographystyle{unsrt}

\begin{thebibliography}{10}

\bibitem{pca}
Matthew Turk and Alex Pentland.
\newblock Eigenfaces for recognition.
\newblock {\em Journal of Cognitive Neuroscience}, 3(1):71--86, January 1991.

\bibitem{lda}
Peter~N. Belhumeur, P.~Hespanha, and David~J. Kriegman.
\newblock Eigenfaces vs. fisherfaces: Recognition using class specific linear
  projection.
\newblock {\em IEEE Transactions on Pattern Analysis and Machine Intelligence},
  pages 711--720, 1997.

\bibitem{lpp}
Xiaofei He and Partha Niyogi.
\newblock Locality preserving projections.
\newblock In {\em Advances in Neural Information Processing Systems (NIPS)}.
  MIT Press, 2003.

\bibitem{lap}
Xiaofei He, Shuicheng Yan, Yuxiao Hu, Partha Niyogi, and Hong jiang Zhang.
\newblock Face recognition using laplacianfaces.
\newblock {\em IEEE Transactions on Pattern Analysis and Machine Intelligence},
  27:328--340, 2005.

\bibitem{glpp}
Huang Sheng, Ahmed Elgammal, Luwen Huangfu, Dan Yang, and Xiaohong Zhang.
\newblock Globality-locality preserving projections for biometric data
  dimensionality reduction.
\newblock In {\em IEEE conference on Computer Vision and Pattern Recognition
  Workshop on Biometrics (CVPRW)}, 2014.

\bibitem{dhlp}
Sheng Huang, Dan Yang, Yongxin Ge, Dengyang Zhao, and Xin Feng.
\newblock Discriminant hyper-laplacian projections with its applications to
  face recognition.
\newblock In {\em IEEE conference on Multimedia and Expo Workshop on HIM
  (ICMEW)}, 2014.

\bibitem{gabor}
Chengjun Liu and Harry Wechsler.
\newblock Gabor feature based classification using the enhanced fisher linear
  discriminant model for face recognition.
\newblock {\em IEEE Transactions on Image Processing}, 11(4):467--476, 2002.

\bibitem{gabor2}
Chengjun Liu and Harry Wechsler.
\newblock Independent component analysis of gabor features for face
  recognition.
\newblock {\em IEEE Transactions on Neural Networks}, 14:919--928, 2003.

\bibitem{mrg}
Yong Xu, Zhengming Li, Jeng-Shyang Pan, and Jing-Yu Yang.
\newblock Face recognition based on fusion of multi-resolution gabor features.
\newblock {\em Neural Computing and Applications}, 23(5):1251--1256, 2013.

\bibitem{grad}
Taiping Zhang, Yuan~Yan Tang, Bin Fang, Zhaowei Shang, and Xiaoyu Liu.
\newblock Face recognition under varying illumination using gradientfaces.
\newblock {\em IEEE Transactions on Image Processing}, 18(11):2599--2606, 2009.

\bibitem{gom}
Ngoc-Son Vu.
\newblock Exploring patterns of gradient orientations and magnitudes for face
  recognition.
\newblock {\em IEEE Transactions on Information Forensics and Security},
  8(2):295--304, 2013.

\bibitem{hog}
Navneet Dalal and Bill Triggs.
\newblock Histograms of oriented gradients for human detection.
\newblock In {\em IEEE Conference on Computer Vision and Pattern Recognition
  (CVPR)}, pages 886--893, 2005.

\bibitem{hogf}
O.~D{\'e}niz, G.~Bueno, J.~Salido, and F.~De~la Torre.
\newblock Face recognition using histograms of oriented gradients.
\newblock {\em Pattern Recognition Letters}, 32(12):1598--1603, September 2011.

\bibitem{part}
Pedro~F. Felzenszwalb, Ross~B. Girshick, David McAllester, and Deva Ramanan.
\newblock Object detection with discriminatively trained part-based models.
\newblock {\em IEEE Transactions on Pattern Analysis and Machine Intelligence},
  32(9):1627--1645, September 2010.

\bibitem{lbp}
Timo Ojala, Matti Pietik\"{a}inen, and Topi M\"{a}enp\"{a}\"{a}.
\newblock Multiresolution gray-scale and rotation invariant texture
  classification with local binary patterns.
\newblock {\em IEEE Transactions on Pattern Analysis and Machine Intelligence},
  24(7):971--987, July 2002.

\bibitem{sift}
David~G. Lowe.
\newblock Distinctive image features from scale-invariant keypoints.
\newblock {\em International Journal of Computer Vision}, 60(2):91--110,
  November 2004.

\bibitem{learn}
Zhimin Cao, Qi~Yin, Xiaoou Tang, and Jian Sun.
\newblock Face recognition with learning-based descriptor.
\newblock In {\em IEEE Conference on Computer Vision and Pattern Recognition
  (CVPR)}, pages 2707--2714. IEEE, 2010.

\bibitem{sadtf}
Rakesh Mehta, Jirui Yuan, and Karen Egiazarian.
\newblock Face recognition using scale-adaptive directional and textural
  features.
\newblock {\em Pattern Recognition}, 47(5):1846--1858, 2014.

\bibitem{lbp2}
Abdenour~Hadid Timo~Ahonen and Matti Pietikainen.
\newblock Face description with local binary patterns: Application to face
  recognition.
\newblock {\em IEEE Transactions on Pattern Analysis and Machine Intelligence},
  28:2037--2041, 2006.

\bibitem{lbp3}
Di~Huang, Caifeng Shan, Mohsen Ardabilian, Yunhong Wang, and Liming Chen.
\newblock Local binary patterns and its application to facial image analysis: A
  survey.
\newblock {\em IEEE Transactions on Systems, Man, and Cybernetics Part C},
  41(6):765--781, November 2011.

\bibitem{fsd}
J~Maatta, A~Hadid, and M~Pietikainen.
\newblock Face spoofing detection from single images using texture and local
  shape analysis.
\newblock {\em IET Biometrics}, 1(1):3--10, 2012.

\bibitem{hogf2}
Alberto Albiol, David Monzo, Antoine Martin, Jorge Sastre, and Antonio Albiol.
\newblock Face recognition using hog-ebgm.
\newblock {\em Pattern Recognition Letters}, 29(10):1537--1543, July 2008.

\bibitem{haar1}
Constantine Papageorgiou and Tomaso Poggio.
\newblock A trainable system for object detection.
\newblock {\em International Journal of Computer Vision}, 38(1):15--33, June
  2000.

\bibitem{haar2}
Rainer Lienhart and Jochen Maydt.
\newblock An extended set of haar-like features for rapid object detection.
\newblock In {\em IEEE International Conference on Image Processing (ICIP)},
  pages 900--903, 2002.

\bibitem{haar3}
Sri-Kaushik Pavani, David Delgado, and Alejandro~F. Frangi.
\newblock Haar-like features with optimally weighted rectangles for rapid
  object detection.
\newblock {\em Pattern Recognition}, 43(1):160--172, January 2010.

\bibitem{haar4}
Paul~A. Viola and Michael~J. Jones.
\newblock Rapid object detection using a boosted cascade of simple features.
\newblock In {\em IEEE Conference on Computer Vision and Pattern Recognition
  (CVPR)}, pages 511--518, 2001.

\bibitem{lsdb}
Yueming Wang, Jianzhuang Liu, and Xiaoou Tang.
\newblock Robust 3d face recognition by local shape difference boosting.
\newblock {\em IEEE Transactions on Pattern Analysis and Machine Intelligence},
  32(10):1858--1870, 2010.

\bibitem{shape3f}
Berk G{\"o}kberk, M~Okan {\.I}rfano{\u{g}}lu, and Lale Akarun.
\newblock 3d shape-based face representation and feature extraction for face
  recognition.
\newblock {\em Image and Vision Computing}, 24(8):857--869, 2006.

\bibitem{shf}
Peijiang Liu, Yunhong Wang, Di~Huang, Zhaoxiang Zhang, and Liming Chen.
\newblock Learning the spherical harmonic features for 3-d face recognition.
\newblock {\em IEEE Transactions on Image Processing}, 22(3):914--925, 2013.

\bibitem{sfc}
Chafik Samir, Anuj Srivastava, and Mohamed Daoudi.
\newblock Three-dimensional face recognition using shapes of facial curves.
\newblock {\em IEEE Transactions on Pattern Analysis and Machine Intelligence},
  28(11):1858--1863, 2006.

\bibitem{ORL}
F.~S. Samaria, F.~S. Samaria, A.C. Harter, and Old Addenbrooke.
\newblock Parameterisation of a stochastic model for human face identification,
  1994.

\bibitem{AR}
Aleix Mart\'{\i}nez and Robert Benavente.
\newblock The ar face database, Jun 1998.

\bibitem{Yaleb}
Athinodoros~S. Georghiades, Peter~N. Belhumeur, and David~J. Kriegman.
\newblock From few to many: Illumination cone models for face recognition under
  variable lighting and pose.
\newblock {\em IEEE Transactions on Pattern Analysis and Machine Intelligence},
  23:643--660, 2001.

\bibitem{lfwa}
Lior Wolf, Tal Hassner, and Yaniv Taigman.
\newblock Similarity scores based on background samples.
\newblock In {\em Asian Conference on Computer Vision (ACCV)}, pages 88--97,
  Berlin, Heidelberg, 2009. Springer-Verlag.

\bibitem{CRC}
D~Zhang, Meng Yang, and Xiangchu Feng.
\newblock Sparse representation or collaborative representation: Which helps
  face recognition?
\newblock In {\em IEEE International Conference on Computer Vision (ICCV)},
  pages 471--478, 2011.

\bibitem{svm}
Chih-Chung Chang and Chih-Jen Lin.
\newblock {LIBSVM}: A library for support vector machines.
\newblock {\em ACM Transactions on Intelligent Systems and Technology}, 2,
  2011.

\bibitem{LR}
Imran Naseem, Roberto Togneri, and Mohammed Bennamoun.
\newblock Linear regression for face recognition.
\newblock {\em IEEE Transactions on Pattern Analysis and Machine Intelligence},
  32(11):2106--2112, 2010.

\bibitem{rcr}
Meng Yang, D~Zhang, and Shenlong Wang.
\newblock Relaxed collaborative representation for pattern classification.
\newblock In {\em IEEE Conference on Computer Vision and Pattern Recognition
  (CVPR)}, pages 2224--2231. IEEE, 2012.

\bibitem{fpatch}
Lin Zhong, Qingshan Liu, Peng Yang, Bo~Liu, Junzhou Huang, and Dimitris~N
  Metaxas.
\newblock Learning active facial patches for expression analysis.
\newblock In {\em IEEE Conference on Computer Vision and Pattern Recognition
  (CVPR)}, pages 2562--2569, 2012.

\bibitem{yupose}
Xiang Yu, Junzhou Huang, Shaoting Zhang, Wang Yan, and Dimitris~N Metaxas.
\newblock Pose-free facial landmark fitting via optimized part mixtures and
  cascaded deformable shape model.
\newblock In {\em IEEE International Conference on Computer Vision (ICCV)},
  2013.

\bibitem{fusion}
Zhen Cui, Wen Li, Dong Xu, Shiguang Shan, and Xilin Chen.
\newblock Fusing robust face region descriptors via multiple metric learning
  for face recognition in the wild.
\newblock In {\em IEEE Conference on Computer Vision and Pattern Recognition
  (CVPR)}, pages 3554--3561, 2013.

\end{thebibliography}

\end{document}